\documentclass[]{TEAI}

\usepackage{float}          %
\usepackage{array}          %
\usepackage{tabularx}       %
\usepackage{makecell}       %
\usepackage{marvosym}       %
\usepackage{pgfplots}       %
\pgfplotsset{compat=1.18}
\DeclareRobustCommand{\Envelope}{\text{\normalfont\Letter}}

\titlespacing\section{0pt}{2.0ex plus 0.4ex minus 0.2ex}{1.0ex plus 0.2ex}
\titlespacing\subsection{0pt}{1.6ex plus 0.3ex minus 0.2ex}{0.8ex plus 0.15ex}
\titlespacing\subsubsection{0pt}{1.2ex plus 0.2ex minus 0.1ex}{0.6ex plus 0.1ex}
\definecolor{erblue}{RGB}{76,114,176}
\definecolor{erorange}{RGB}{221,132,82}
\definecolor{ergreen}{RGB}{85,168,104}
\definecolor{erred}{RGB}{196,78,82}
\definecolor{ercup}{HTML}{4874CB}
\definecolor{ertable}{HTML}{EE822F}
\definecolor{erdisk}{HTML}{f2ba02}
\definecolor{eravg}{HTML}{75BD42}
\newcommand{\RobotSuccess}[1]{Robot success~(#1)}
\newcommand{\RobotRecovery}[1]{Robot recovery~(#1)}
\newcommand{\HumanSuccess}[1]{Human success~(#1)}
\newcommand{\HumanRecovery}[1]{Human recovery~(#1)}

\newcommand{\Beq}{B_{\mathrm{RobotEq}}}

\title{EgoRecovery: Acquiring Failure Recovery Ability Through Human Recovery Demonstration}

\author[1,2,*]{Zuhao Ge}
\author[1,2,3,*]{Yuchen Zhou}
\author[3,\dagger]{Weitao Zhou}
\author[3]{Minglei Li}
\author[1,2]{Xinyu Li}
\author[1,2]{Chao Wu}
\author[1,2]{Hanwen Zhao}
\author[1,2]{Haotian Wang}
\author[1,2]{Zuxuan Wu}
\author[1,2,\Envelope]{Xiaosong Jia}
\author[1,2]{Yu-Gang Jiang}

\affiliation[1]{Institute of Trustworthy Embodied AI (TEAI), Fudan University}
\affiliation[2]{Shanghai Key Laboratory of Multimodal Embodied AI}
\affiliation[3]{Simple AI}
\renewcommand\affiliationlist{%
  {\affiliationfont
  $^{1}$Institute of Trustworthy Embodied AI (TEAI), Fudan University\par
  $^{2}$Shanghai Key Laboratory of Multimodal Embodied AI, $^{3}$Simple AI\par}%
}

\contribution[*]{Equal contribution}
\contribution[\dagger]{Project Leader}
\contribution[\Envelope]{Corresponding author}
\checkdata[Code]{\url{https://github.com/EgoRecovery/EgoRecovery}}

\abstract{Robust embodied robots should be able to recover from failures and
retry tasks in order to operate reliably in unstructured and noisy real-world
environments. Achieving this capability requires training policies on data that
captures recovery behaviors. However, collecting such data through robot
teleoperation is difficult to scale, as it is time-consuming to induce diverse
failure states, perform corrective actions, and reset the environment. This
challenge is further exacerbated by the high diversity of failure modes, which
demands substantially more recovery data than success demonstrations. In this
work, we show that egocentric human data capturing failure recovery processes
provides a scalable alternative. By efficiently arranging task-level failure
configurations and recording short recovery segments, human operators can
generate more than $10\times$ as much valid recovery data per hour compared to
robot teleoperation under our protocol. To address the embodiment gap between
human and robot, we propose \textsc{EgoRecovery}, a co-training framework for
learning recovery behavior, where human recovery demonstrations are aligned to
a compact corrective-intent space shared with robot data, which captures the
timing and magnitude of correction. Only a small number of robot recovery
demonstrations are required to connect this intent to executable robot actions.
At deployment, a learned recovery gate predicts when correction is needed from
robot observations and activates the corrective intent only in recovery states.
Experiments on real-world recovery tasks show that \textsc{EgoRecovery}
improves success from failure starts over robot-only recovery, direct
co-training with human recovery data, and direct intent-transfer baselines.}

\begin{document}

\maketitle
\begingroup
\renewcommand{\thefootnote}{\fnsymbol{footnote}}
\footnotetext[1]{Work is done when Yuchen interns at Simple AI.}
\endgroup

\begin{figure}[H]
  \centering
  \includegraphics[width=0.95\textwidth]{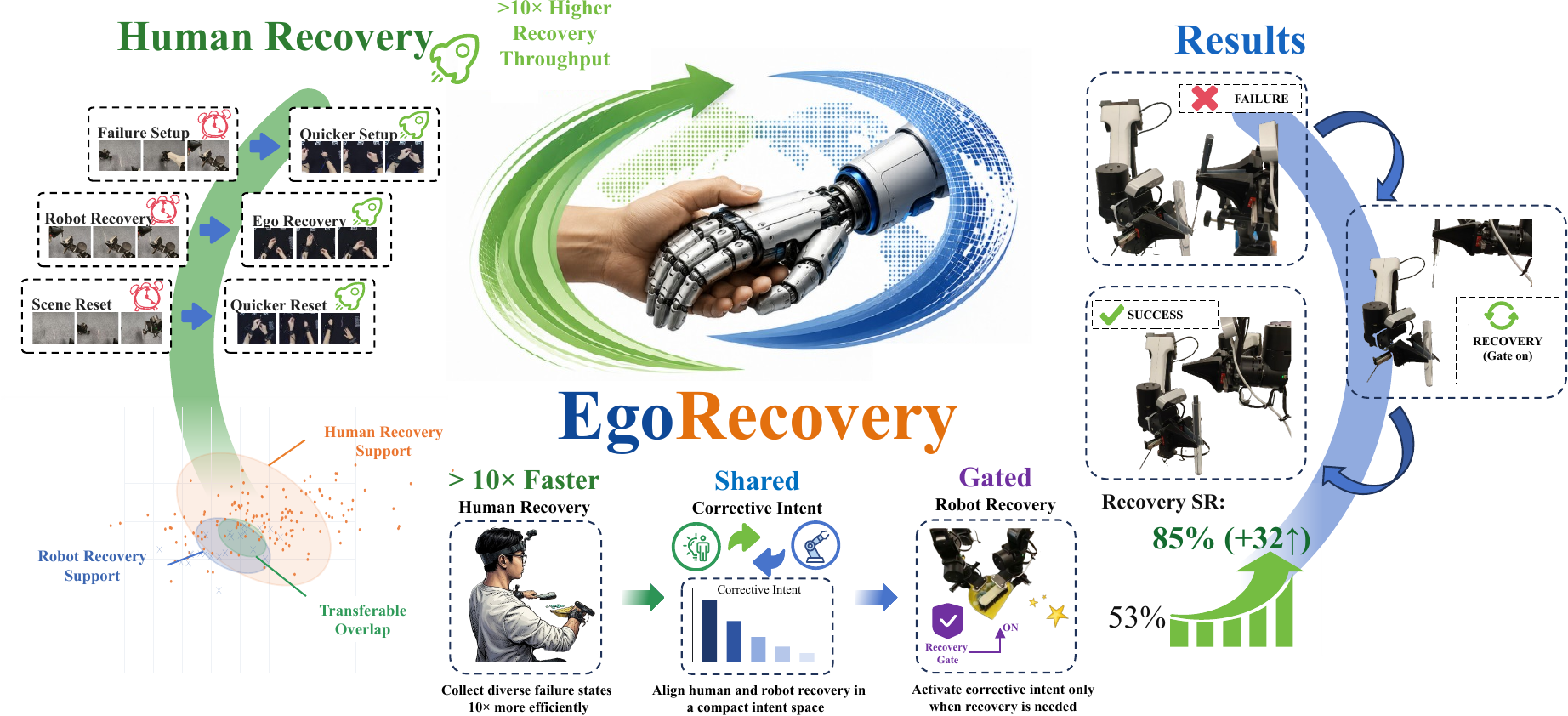}
  \caption{\textbf{Overview of \textsc{EgoRecovery}.} Egocentric human
  recovery data collection yields about $10\times$ more data and
  broader coverage of failure states than robot teleoperation collection, while \textsc{EgoRecovery} transfers this
  signal through a gated corrective-intent bottleneck.}
  \label{fig:teaser}
  \vspace{-0.7em}
\end{figure}

\vspace{-0.6em}
\teaiabstract
\clearpage

\section{Introduction}
Robust embodied robots require policies that can recover from their own mistakes. Policies trained solely on successful demonstrations perform well from clean initial states but degrade when closed-loop rollouts drift away from the demonstrated manifold, a well-known instance of covariate shift in imitation learning~\cite{dagger2011,dart2017}. Recent methods demonstrate that explicit recovery and correction data can substantially improve long-horizon robustness~\cite{rac2025,sop2026}. However, collecting such data is costly. Unlike successful demonstrations, acquiring a recovery trajectory often requires repeatedly inducing or reaching a meaningful failure state, verifying it, teleoperating a corrective action, and resetting the hardware. Moreover, because failure modes can be highly diverse, covering the distribution of possible off-manifold states demands many varied trajectories, further increasing the data-collection cost of learning recovery behaviors.

Recently, learning from egocentric human data has shown strong potential to improve the generalization of robot policies~\citep{bahl2022human2robot,mimicplay2023,egomimic2024,egobridge2025,
zhu2025emma,innon2025,egoverse2026,egoscale2026}, yet they use human data primarily for nominal
task execution or broad pretraining. In this work, \textbf{we exploit the speed advantage of human egocentric data collection to gather failure-recovery data as efficient coverage of diverse failure modes}. A human demonstrator can collect such data quickly by setting up task-level failure configurations, recording only short recovery segments, and moving rapidly to the next failure mode. In our setting, measured as valid recovery segments per operator hour under the collection protocol summarized in Table~\ref{tab:throughput_audit} (full protocol in Appendix~\ref{appx:human_collection}), this yields $10\times$ more data than robot teleoperation collection.

Deciding what to share across embodiments is a central challenge in
cross-embodiment imitation~\citep{egobridge2025,lidea2026,bridgeact2026,mothra2026}. Human hands and robot grippers differ in
kinematics, contact mechanics, control frequency, and action
representation. Sharing the action decoding head can bias the robot toward actions that do not match its embodiment, while simply adopting a different decoding head might not fully utilize the information in human recovery data. Thus, we propose to share \emph{corrective intent} rather than full human motion across embodiments. Corrective intent describes the timing and magnitude of a short correction within a recovery segment. This choice follows cross-embodiment imitation methods that share compact behavior descriptions rather than copy trajectories~\citep{xskill2023,uniskill2025,mint2026}. It is also well matched for learning recovery behaviors, since a recovery segment starts from a failed configuration and aims to return the task to a feasible state. Human recovery data can indicate when to retreat, realign, or retry, and how much adjustment is needed. The robot policy still learns direction, contact, and executable actions from robot data.

Specifically, we propose \textsc{EgoRecovery}, a joint training method that maps human recovery supervision to robot recovery through corrective intent. It shares visual backbones across embodiments and keeps embodiment-specific action heads, following HPT style co-training~\citep{hpt2024}. A shared recovery intent head is designed to extract corrective intent from human and robot recovery segments, which aligns two domains in a shared yet compressed intent representation space. Moreover, a recovery gate head is trained to predict whether a recovery behavior is needed based on current observations and states, supervised by recovery and success phase labels. The predicted gate modulates features in the robot action head, allowing corrective intent to affect recovery states while preserving nominal execution learned from robot data. %

In summary, \textbf{our contributions are as follows:}
\vspace{-0.35em}
\begingroup
\setlength{\leftmargini}{1.15em}
\begin{itemize}
  \setlength{\itemsep}{0.05em}
  \setlength{\parsep}{0pt}
  \setlength{\parskip}{0pt}
  \setlength{\topsep}{0.05em}
  \setlength{\partopsep}{0pt}
  \item We introduce egocentric human recovery as a scalable source of recovery
  supervision that covers diverse task-level failure states and provides more
  than $10\times$ as much valid recovery supervision per operator hour as robot
  teleoperation under our protocol.
  \item We propose \textsc{EgoRecovery}, which transfers human recovery through
  corrective intent rather than human action labels, with a recovery intent head
  for shared intent and a recovery gate head for deciding when it should affect
  robot actions.
  \item We evaluate real robot recovery under controlled data budgets and show
  that \textsc{EgoRecovery} improves closed-loop recovery over robot-only
  recovery, human success co-training, direct co-training with human recovery data, and
  ungated intent transfer baselines.
\end{itemize}
\endgroup

\section{Related Work}

\noindent\textbf{Recovery and correction in imitation learning.}\par
\noindent
Recovery supervision addresses the covariate shift that arises when imitation policies leave the demonstration manifold. DAgger collects labels on states induced by the policy, and DART trains policies to handle perturbed states around demonstrations \citep{dagger2011,dart2017,mandlekar2021what,hgdagger2019,thriftydagger2021}. Recent robot methods extend this idea by collecting recovery trajectories, using intervention signals,
detecting failures at runtime, or generating corrective behavior with learned models
\citep{rac2025,sop2026,liu2023runtime,rewindil2026,flowcorrect2026,tamen2026,hiwm2026,wmdagger2026,reflect2023}.
These works show that supervision near failure states can improve closed-loop robustness, but diverse failure-state supervision remains costly when corrective data is obtained from robot rollouts, robot interventions, or surrogate robot trajectories. \textsc{EgoRecovery} keeps recovery supervision as the target but moves part of failure state acquisition to egocentric human recovery segments
staged around task-level failures.

\vspace{0.25em}
\noindent\textbf{Egocentric human data for robot learning.}\par
\noindent
Egocentric human data is attractive because humans can generate diverse interactions without operating robot hardware. Prior work uses human videos and human play to provide learning signals for robot policies \citep{bahl2022human2robot,mimicplay2023,hoipretrain2024,zeromimic2025,egozero2025,ma2026humanvideosurvey,r3m2022}. More recent systems co-train egocentric human demonstrations with robot data, align human and robot representations, or use heterogeneous backbones that share features while preserving robot action heads
\citep{egomimic2024,egobridge2025,hpt2024,lidea2026,innon2025,easymimic2026,warped2026,unidex2026,maniptrans2025,lin2026systematic,cotrain_mechanism2026,xrzerog02026,libravla2026,openx2024,umi2024,diffusionpolicy2023}.
Other recent work explores human gaze and active vision cues from egocentric data for manipulation~\citep{gazevla2026,activeglasses2026}.
Large egocentric datasets further show that human data can scale across tasks and environments \citep{ego4d2022,egoverse2026,egolive2026,zhu2025emma,egoscale2026,egodex2025,aoe2026,mimicgen2023}, and vision-language-action models pretrained on such corpora extend this scaling to control~\citep{vitra2025,vipavla2025,egovla2025,hrdt2025,beingh052026,dreamdojo2026}.
These works mainly use human data for nominal execution, broad pretraining, or action alignment. Our direct human recovery mix applies this co-training setup to recovery data, while \textsc{EgoRecovery} routes the same type of human recovery data through corrective intent rather than direct robot action labels.

\vspace{0.25em}
\noindent\textbf{Transferable representations across embodiments.}\par
\noindent
Human recovery segments cannot be used as robot demonstrations unless shared task information is separated from embodiment-specific execution. Cross-embodiment imitation therefore often transfers compact behavior descriptions instead of raw trajectories, including latent plans, skills, affordances, and intent variables
\citep{playlmp2020,xskill2023,uniskill2025,bridgeact2026,mothra2026,mint2026,clap2026,twohandedafforder2025,conla2026,jala2026,mvplam2026,unilact2026,care2026,anchorrefine2026,resvla2026,vlajepa2026,fast2025}.
These abstractions motivate our use of corrective intent, but the scope here is more specific. \textsc{EgoRecovery} does not aim to learn a generic human intention prior. It transfers a compact corrective-intent target, while robot observations and robot data determine direction, contact, and executable actions. The recovery gate further restricts this transfer to states where correction is needed, which separates our setting from direct human recovery mixing and ungated intent transfer.
\vspace{0.25em}

\section{Method}
\label{sec:method}

\begin{figure}[t]
  \centering
  \includegraphics[width=\linewidth]{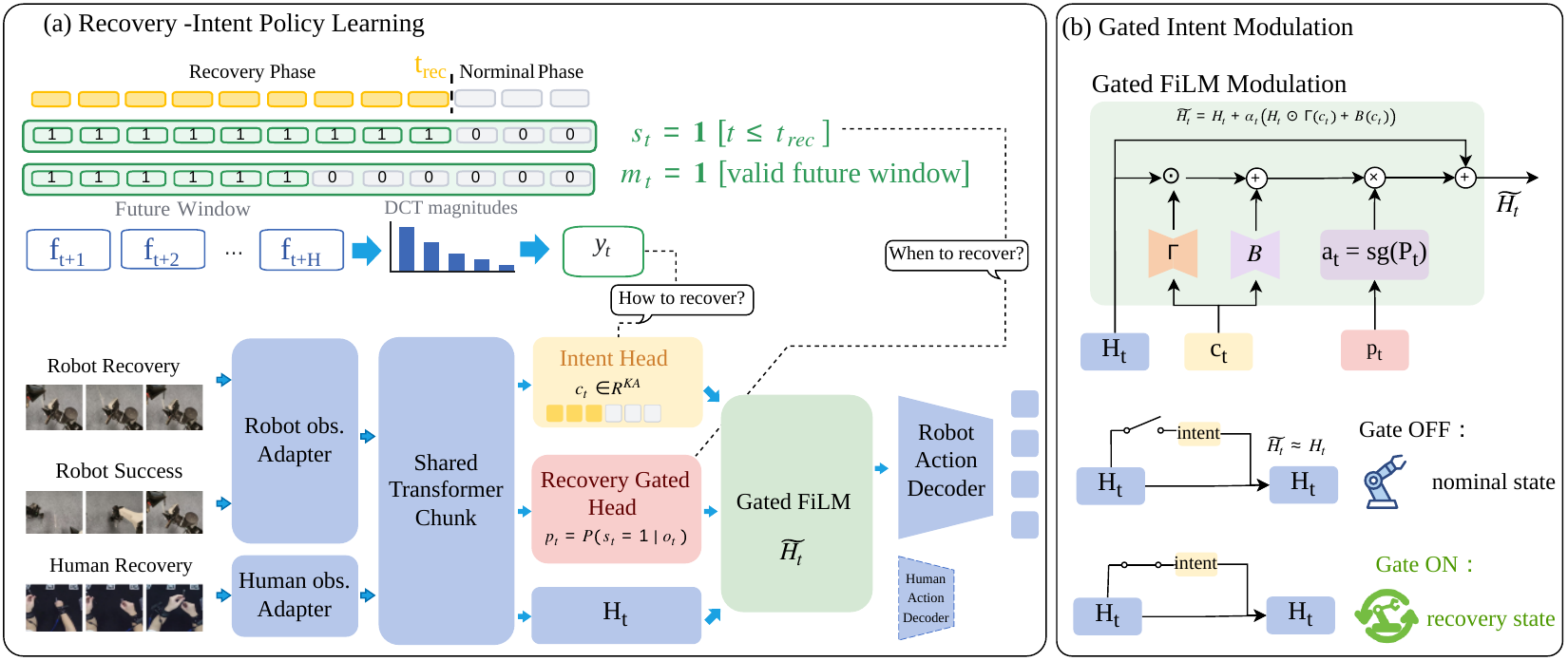}
  \vspace{-0.75em}
  \caption{\textbf{\textsc{EgoRecovery} method.}
  \textbf{a,} During training, robot success preserves nominal execution, robot
  recovery grounds corrective intent in executable robot actions, and human
  recovery broadens failure-state coverage. Recovery annotations define the
  recovery-state label $s_t$ and the intent mask $m_t$, which supervise intent
  prediction and recovery-gate prediction. \textbf{b,} At deployment, the robot
  uses only robot observations. The predicted intent $c_t$ modulates decoder
  features through gated FiLM only when the predicted gate $p_t$ indicates that
  recovery is needed.}
  \label{fig:pipeline}
  \vspace{-0.75em}
\end{figure}

\textsc{EgoRecovery} is motivated by the challenge of learning recovery behaviors.  Robot recovery data
provides executable corrective action labels, but collecting it across many
failure states is expensive. Human recovery data scales better due to the fast collection speed, but human motion cannot be directly used as robot action supervision. In \textsc{EgoRecovery}, we
use human recovery to broaden coverage of failure states and design co-training structure to enable cross-embodiment recovery behavior learning.
Figure~\ref{fig:pipeline} gives an overview of the proposed method. Section~\ref{sec:data} defines the data source, 
Section~\ref{sec:factor} describes how to extract the
compact and shared corrective intent, and Section~\ref{sec:align} explains
how the intent is used to decode robot action.

\subsection{Human and Robot Data Collection}
\label{sec:data}
Robot recovery data provides executable corrective actions, but it is expensive to
scale because each episode requires staging a failure state,
checking that it is valid, teleoperating the correction, and resetting the
hardware. Collecting egocentric human recovery data offers a much cheaper alternative. As shown in Table~\ref{tab:throughput_audit}, under the same protocol, including
staging, reset, synchronization, and discarded attempts, human recovery provides
about $10\times$ more accepted recovery episodes per operator hour than teleoperated robot
recovery.

\noindent\textbf{Data types.} For each task, there are four data types explored in this paper: robot success, robot recovery, human success, and human recovery. 
Robot success data provides nominal robot action labels to successfully execute the task while Robot
recovery data provides failed robot start states and corrective actions. 
Human success and recovery data are similar to their robot counterparts but are not directly used to train the robot action head.
They are only used to extract the compact intent and supervise the human action decoding head. We denote the number of episodes per type as follows: \RobotSuccess{50}, \RobotRecovery{50}, \HumanSuccess{300}, and \HumanRecovery{300}. See Appendix~\ref{appx:human_collection} for the collection protocol and annotation details.

\noindent\textbf{Recovery labels.}  Each recovery episode starts from a failure state and is divided into two phases: recovery and ordinary execution. 
The annotated temporal boundary $t_{\mathrm{rec}}$ separates the two phases and indicates that corrective motion has ended and ordinary task execution resumes. The recovery state label $s_t$ is set as true
until time-step $t_{\mathrm{rec}}$ and after that as false, which is used to supervise a Recovery Gate Head (detailed in Sec.~\ref{sec:align}) to recognize whether
recovery behavior is required. The intent mask $m_t$ is defined for frames whose future
motion window stays inside the recovery portion and passes checks for padding,
truncation, missing motion, and near-stationary motion. Success episodes set
both $s_t$ and $m_t$ as False, providing negative examples for the gate and preventing
ordinary execution from defining corrective intent.

\subsection{Corrective Intent for Human to Robot Recovery Learning}
\label{sec:factor}
Human recovery can guide robot recovery only through information that remains
meaningful across embodiments. Human hands and robot grippers differ in contact,
kinematics, and control, so copying human motion would turn embodiment-specific
motion into an incorrect robot action target. We therefore use corrective
intent as an intermediate supervision target rather than treating human motion
as robot action. In this paper, corrective intent means a coarse envelope of the
remaining recovery motion. It describes how much correction remains over a short
future window, while the gate $p_t$ separately predicts whether the current
observation is a recovery state. The intent leaves direction, contact, and
executable motor commands to robot observations and robot action supervision.

We construct this target from short future end-effector motion windows rather
than manual semantic labels or full 14D action chunks. For a frame selected by
$m_t$, let $P_{t,a}^{d}\in\mathbb{R}^{L\times 3}$ be the future active
end-effector positions for embodiment $d$ and active effector $a$, and let
$s_d$ be the embodiment scale. To make the envelope compact and less sensitive
to small timing or contact differences, we project the normalized future motion
onto a low-frequency discrete cosine transform (DCT) basis:
\begin{equation}
\Delta P_{t,a}^{d}[\ell]
=
\frac{P_{t,a}^{d}[\ell]-P_{t,a}^{d}[0]}{s_d},
\qquad
y_t^{d}
=
\Bigl[
\bigl\|
\left(C_K \Delta P_{t,a}^{d}\right)_{k,:}
\bigr\|_2
\Bigr]_{a=1{:}A,\;k=0{:}K-1}
\in \mathbb{R}^{KA}.
\label{eq:intent_target}
\end{equation}
Here $C_K$ contains the first $K$ low-frequency DCT basis vectors over the
future window. The DCT is used only to summarize the temporal envelope; the
spatial norm removes direction-specific displacement and leaves a magnitude
profile for each active effector. Thus $y_t$ is not an action plan or semantic
intent label; it only specifies how the remaining correction magnitude unfolds
over the window. This makes the target transferable but intentionally
incomplete: executable direction and contact are learned from robot observations
and robot recovery actions. In the main experiments, one active effector and
four temporal coefficients give a 4D target.

The intent mask determines where this target can supervise the policy. Applying
the intent loss only when $m_t$ is true prevents nominal motion or invalid
future segments from becoming examples of corrective intent. This gives the two
recovery sources complementary roles. Human recovery trains the intent predictor
on many related failure states, while robot recovery supplies the same target
together with robot action labels. The robot decoder can then learn how the
predicted intent should be realized as executable control.

\subsection{Gated Intent Modulation for Robot Recovery Execution}
\label{sec:align}

The predicted intent should affect robot action decoding only when correction is
needed. Otherwise, if the decoder receives the intent at every time step, recovery
supervision can interfere with behavior learned from robot success data. We
therefore train a recovery gate that predicts from the current observation whether
$c_t$ should modulate the robot decoder.

The policy uses an HPT style heterogeneous backbone~\citep{hpt2024} with
embodiment-specific adapters and decoders around a shared trunk. Each
observation is encoded by its embodiment adapter and processed by the shared
trunk. A pooled representation feeds two shared prediction heads. The Intent
Head predicts $c_t\in\mathbb{R}^{KA}$, and the Recovery Gate Head predicts
$p_t$, the probability that the observation requires recovery behavior. The
robot decoder maps robot features to robot actions, while an auxiliary human
decoder maps human features to human motion. The auxiliary human motion loss is
used only for the human-domain decoder and shared representation; it never
supervises the robot decoder and is not treated as a cross-embodiment action
label. This separates embodiment-specific action spaces while allowing recovery
observations from both embodiments to train the shared trunk and intent head.

The robot decoder receives $c_t$ through a gated residual FiLM modulation before
the output projection. Let $H_t\in\mathbb{R}^{D}$ be the robot decoder feature
before this projection, and let
$\Gamma,B:\mathbb{R}^{KA}\rightarrow\mathbb{R}^{D}$ map $c_t$ to feature-wise
scale and shift terms. We set the scalar gate
$\alpha_t=\mathrm{sg}(p_t)$, which broadcasts over feature dimensions and stops
gradients through the gate in the modulation path. The residual form makes the
identity path explicit:
\begin{equation}
\setlength{\abovedisplayskip}{0pt}
\setlength{\belowdisplayskip}{0pt}
\setlength{\abovedisplayshortskip}{0pt}
\setlength{\belowdisplayshortskip}{0pt}
\widetilde{H}_t
=
H_t + \alpha_t\left(H_t \odot \Gamma(c_t) + B(c_t)\right),
\qquad
\hat{a}_t = W_r(\widetilde{H}_t).
\label{eq:gated_modulation}
\end{equation}
When $p_t$ is near zero, $\widetilde{H}_t$ reduces to the base decoder feature.
When $p_t$ is high, the residual term injects corrective intent before action
prediction. On selected robot recovery frames, the same observation contributes
an intent loss on $c_t$ and a robot action loss through the modulated decoder.
The robot action loss is allowed to back-propagate through $\Gamma(c_t)$ and
$B(c_t)$ into the Intent Head, so robot recovery can teach how the predicted
intent affects executable actions. The masked intent loss anchors $c_t$ to the
shared target $y_t$, preventing it from drifting into an unconstrained
robot-only latent. The stop gradient keeps the gate supervised by recovery state
labels rather than optimized indirectly through imitation loss.

The training objective follows the same separation of responsibility across data
sources:

\begin{equation}
\setlength{\abovedisplayskip}{0pt}
\setlength{\belowdisplayskip}{0pt}
\setlength{\abovedisplayshortskip}{0pt}
\setlength{\belowdisplayshortskip}{0pt}
\mathcal{L}
=
\mathcal{L}_{\mathrm{bc}}^r
+ \mathcal{L}_{\mathrm{bc}}^h
+ \lambda_c\,\mathcal{L}_{c}
+ \lambda_g\,\mathcal{L}_{g}
+ \lambda_n\,\mathcal{L}_{n}.
\label{eq:training_objective}
\end{equation}

Here $\mathcal{L}_{\mathrm{bc}}^r$ is behavior cloning on robot success and
robot recovery samples, and $\mathcal{L}_{\mathrm{bc}}^h$ is behavior cloning for
the auxiliary human decoder only. The intent loss
$\mathcal{L}_{c}=\mathbb{E}_{m_t=1}[\|c_t-y_t\|_1]$ anchors the predicted intent
to the shared target on valid recovery windows. The gate loss
$\mathcal{L}_{g}=\mathbb{E}[\mathrm{BCE}(p_t,s_t)]$ trains recovery-state
detection. The nominal regularizer
$\mathcal{L}_{n}=\mathbb{E}_{s_t=0}[p_t(\|\Gamma(c_t)\|_2^2+\|B(c_t)\|_2^2)]$
penalizes nonzero intent modulation on non-recovery frames. In this objective,
human recovery shapes the intent and gate, robot recovery connects the predicted
intent to executable robot actions, and robot success data preserves nominal
execution.

At deployment, the robot uses only robot observations, the predicted intent, the
predicted gate, and the robot decoder. Recovery labels and intent targets are
used only during training. Appendix~\ref{appx:intent_target} and
Appendix~\ref{appx:training} give implementation details for the target
construction, window length, scale computation, sampling ratios, and loss
weights.

\section{Experiments}
\label{sec:experiments}

\begin{figure}[!t]
\centering
\includegraphics[width=\textwidth]{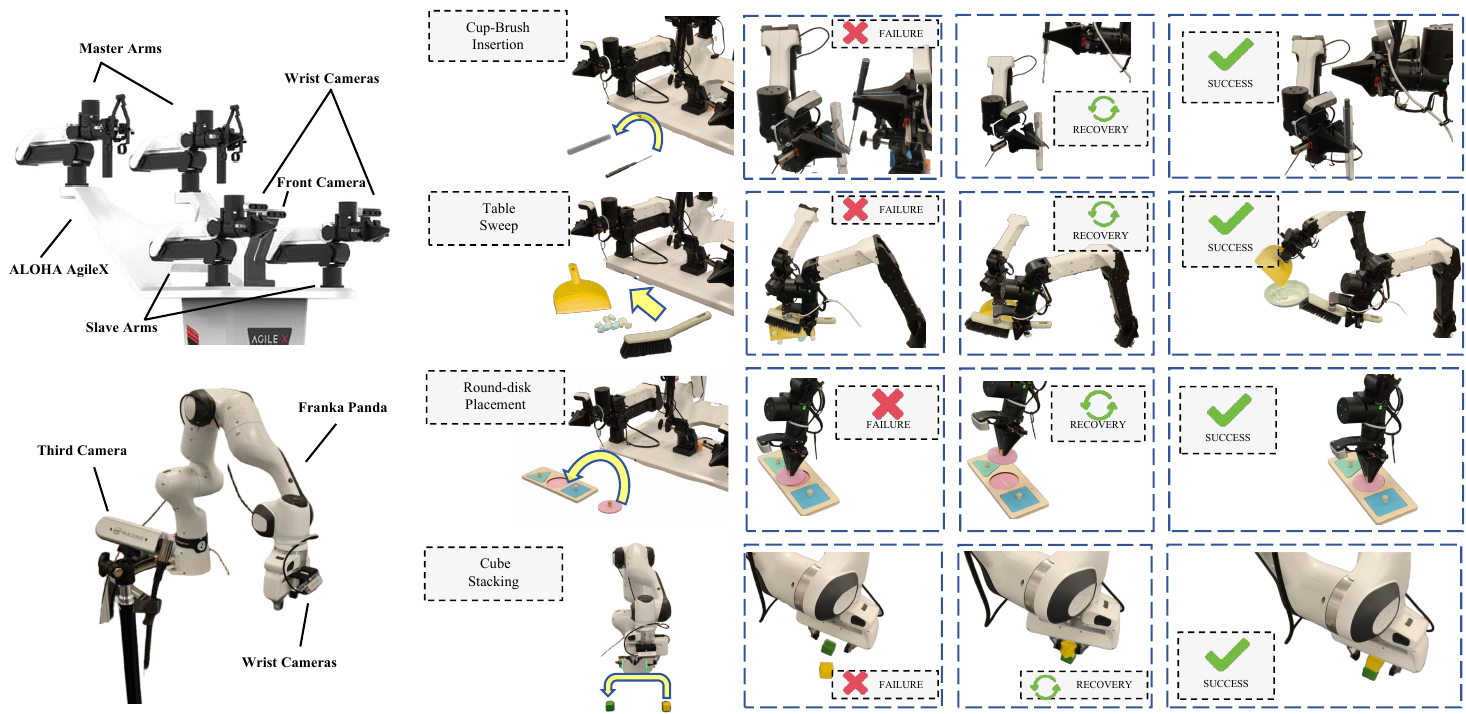}
\vspace{-0.55em}
\caption{\textbf{Experimental platforms and task objects.} Real robot tabletop
settings used for closed-loop evaluation across four manipulation tasks: cup
brush insertion, table sweep, round disk placement, and cube stacking.}
\label{fig:setup}
\vspace{-0.75em}
\end{figure}

\subsection{Experimental Setup}

  \noindent\textbf{Protocol.} All evaluations are conducted in closed-loop rollouts on
  real robots. At test time, policies receive only robot observations. Human
  videos, recovery window labels, and corrective intent targets are not
  available. We report two success metrics that separate ordinary task
  execution from recovery behavior.
  \emph{Initial SR} measures task completion from the normal initial state.
  \emph{Recovery SR} measures task completion from the off-nominal state.
  A rollout is counted as successful if the robot completes the task within the
  prescribed time budget. Full rollout rules, timeouts, success criteria, and
  the egocentric recording setup used for human recovery collection are provided
  in the appendix; Appendix Figure~\ref{fig:hand_setting} shows the recording
  setup. Figure~\ref{fig:setup} shows the robot platforms and task objects used
  for closed-loop evaluation.

  \noindent\textbf{Tasks.} We evaluate four tabletop manipulation tasks: cup
  brush insertion, table sweep, round disk placement, and cube stacking. They
  span retreat-and-realign, corrective sweeping, lift-realign-place, and stack
  stabilization patterns. Failure starts are constructed within the same
  recovery pattern as the nominal task, so the evaluation tests recovery from
  related failure states rather than transfer to unrelated skills.

  \textbf{Baselines.} We name data sources directly by embodiment and phase.
  \RobotSuccess{n} denotes $n$ robot success demonstrations, \RobotRecovery{n}
  denotes $n$ robot recovery demonstrations, \HumanSuccess{n} denotes $n$ human
  success demonstrations, and \HumanRecovery{n} denotes $n$ human recovery
  demonstrations. We also report a robot-equivalent recovery budget, denoted
  by $\Beq$, in which each human recovery demonstration counts as one tenth of a
  robot recovery demonstration based on the measured 10$\times$ collection-rate
  advantage in Table~\ref{tab:throughput_audit}.
  The direct human recovery mix is the co-training baseline with matched data. It
  uses the same data and backbone as \textsc{EgoRecovery}, but removes the
  corrective intent target, Recovery Gate Head, and gated modulation. Robot-only
  recovery rows control for recovery data collected on
  the robot. We use the \RobotRecovery{50} row and the cost matched
  \RobotRecovery{80} reference in Figure~\ref{fig:budget} as robot-only
  references, rather than as direct comparisons to alternative robot
  recovery data collection algorithms.

\subsection{Main Results}
  \label{sec:exp:main}

\begin{figure}[t]
\centering
\includegraphics[width=\textwidth]{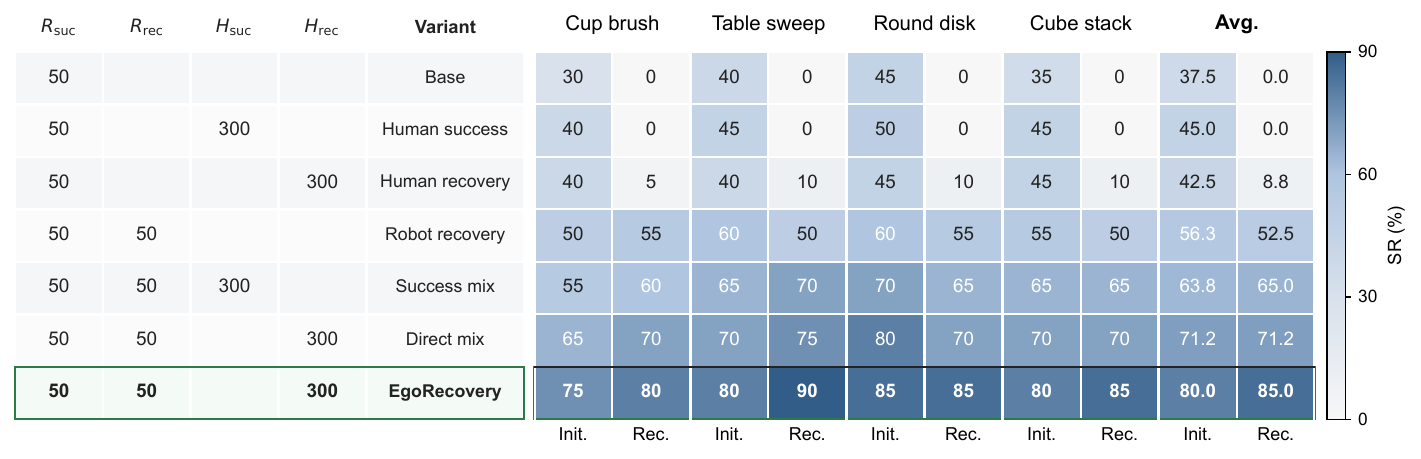}
\vspace{-6mm}
  \caption{\textbf{Main closed-loop results across four tabletop tasks.}
  Each task group reports Initial SR and Recovery SR (\%) for cup brush,
  table sweep, round disk, cube stack, and their average. In the method labels,
  $R_{\mathrm{suc}}$, $R_{\mathrm{rec}}$, $H_{\mathrm{suc}}$, and
  $H_{\mathrm{rec}}$ denote robot success, robot recovery, human success, and
  human recovery demonstration counts.}
\label{fig:main_results}
\vspace{-0.75em}
\end{figure}

  Figure~\ref{fig:main_results} reports the main closed-loop comparison.
  This evaluation uses common failure starts. For each task, the failure
  configuration is randomly sampled from the task's standard off-nominal range.
  These starts come from the same recovery family as the robot recovery data,
  but use varied failed configurations rather than reusing the exact robot
  recovery training starts.

  Each method is tested with 20 rollouts per task and start-state regime on the
  same fixed initial states and recovery starts; Appendix~\ref{appx:additional_results}
  reports the corresponding counts and Wilson 95\% confidence intervals.

\textbf{Nominal demonstrations do not solve recovery.}
Methods trained only on nominal demonstrations did not solve the recovery
benchmark. \RobotSuccess{50} and \RobotSuccess{50} + \HumanSuccess{300} both
achieved 0.0 average Recovery SR, whereas adding \RobotRecovery{50} raised
average Recovery SR to 52.5. This result shows that acting from failure states
requires explicit recovery supervision rather than additional nominal
co-training alone.

\textbf{Human recovery is useful, but it is not sufficient without robot recovery grounding.}
Human recovery alone provided only limited benefit:
\RobotSuccess{50} + \HumanRecovery{300} reached 8.8 average Recovery SR. When
human recovery was added to the robot anchor
\RobotSuccess{50} + \RobotRecovery{50}, performance improved substantially: the
direct human recovery mix reached 71.2 average Recovery SR, and
\textsc{EgoRecovery} further improved this to 85.0 while also achieving the
highest average Initial SR of 80.0. These comparisons indicate that human
recovery provides useful corrective-state supervision, but that robot recovery
remains necessary to ground this signal in executable robot actions.

\textbf{The gain is explained by gated corrective-intent transfer.}
The key controlled comparison is between the direct human recovery mix and
\textsc{EgoRecovery}. Both use the same data composition
(\RobotSuccess{50}+\RobotRecovery{50}+\HumanRecovery{300}) and the same shared
backbone, but only \textsc{EgoRecovery} routes human recovery through the
corrective-intent bottleneck and Recovery Gate Head. This design raises average
Recovery SR from 71.2 to 85.0, showing that the improvement is not explained by
adding human recovery data alone. Instead, human recovery is most effective
when it is transferred through a recovery-specific, gated pathway that
influences robot action decoding only when correction is needed.

\subsection{Data Efficiency}
\label{sec:exp:efficiency}

We next examine whether egocentric human recovery provides an efficient route
to improved recovery. We do so with three pieces of evidence: scaling with
human data volume, cost-matched substitution between robot and human recovery,
and an operator-hour throughput audit.

\begin{figure}[t]
\centering
\includegraphics[width=\textwidth]{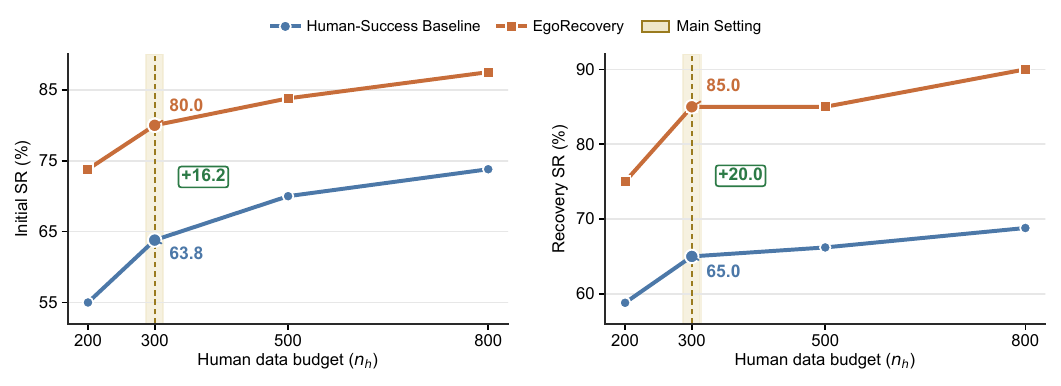}
\vspace{-6mm}
  \caption{\textbf{Human data scaling under fixed robot recovery grounding.}
  \textbf{a,} Average Initial SR. \textbf{b,} Average Recovery SR. The human
  data budget varies from 200 to 800 episodes, while all settings use
  \RobotSuccess{50} and \RobotRecovery{50}. The comparison contrasts the
  Human-Success Baseline and \textsc{EgoRecovery}, which replaces nominal
  human data with \HumanRecovery{n} routed through the proposed
  recovery-specific pathway. The shaded band marks the Main Setting.}
\label{fig:scaling}
\vspace{-0.75em}
\end{figure}

Figure~\ref{fig:scaling} first asks whether the gain can be explained by adding
more human data under fixed robot recovery grounding. All rows use
\RobotSuccess{50} and \RobotRecovery{50}; the left block adds
\HumanSuccess{n}, and the right block replaces it with \HumanRecovery{n}
through \textsc{EgoRecovery}.

\textbf{Human recovery scaling improves recovery more directly than nominal
human co-training.}
Adding nominal human success data after robot recovery grounding improves
average Initial SR from 55.0 to 73.8 as the human budget grows from 200 to 800
episodes. Its effect on Recovery SR is more limited. In contrast,
\textsc{EgoRecovery} achieves higher Recovery SR at every matched human budget.
It reaches 75.0 at 200 episodes, 85.0 at the main 300 episode setting, and 90.0 at 800
episodes. This pattern shows that additional human data help recovery most when
they are collected from failure states and transferred through the
recovery-specific pathway, rather than added only as nominal execution data.

\begin{figure}[t]
\centering
\includegraphics[width=0.96\textwidth]{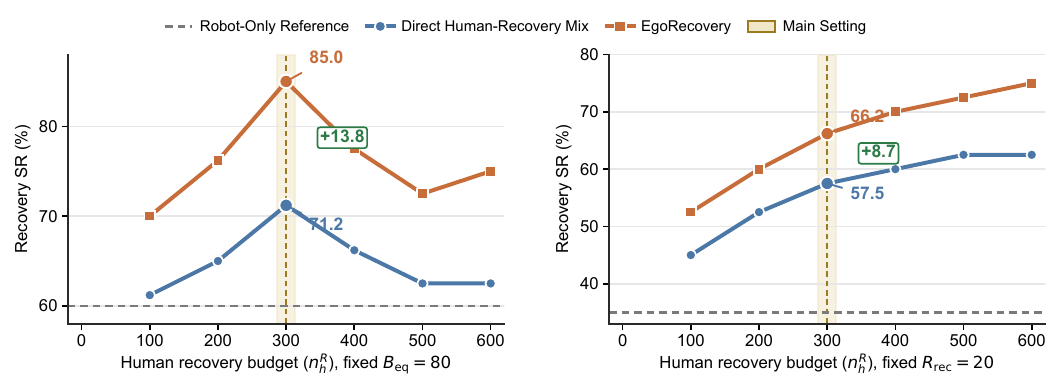}
\vspace{-4mm}
\caption{\textbf{Recovery budget comparison.} \textbf{a,} Average Recovery SR
under cost-matched substitution, where the robot-equivalent recovery budget is
fixed at $\Beq=80$ while the amount of human recovery data varies.
\textbf{b,} Average Recovery SR under low robot grounding, where
\RobotRecovery{20} is fixed while the human recovery budget varies. The
comparison contrasts the Robot-Only Reference, the Direct Human-Recovery Mix,
and \textsc{EgoRecovery}. The shaded band marks the Main Setting.}
\label{fig:budget}
\vspace{-0.45em}
\end{figure}

Figure~\ref{fig:budget} evaluates two complementary budget regimes.
Panel (a) keeps $\Beq$ fixed using the measured
10$\times$ human to robot recovery conversion, and Panel (b) fixes a low
\RobotRecovery{20} grounding budget while increasing human recovery.

\textbf{A cost matched tradeoff keeps a nontrivial robot recovery
anchor.}
Under the fixed robot equivalent recovery budget $\Beq=80$ in
Figure~\ref{fig:budget} (a), moderate substitution of robot recovery with human
recovery outperformed the robot-only reference.
\RobotRecovery{50} with \HumanRecovery{300} reaches 85.0 average Recovery SR,
compared with 60.0 for the \RobotRecovery{80} reference. The direct mix also
improves with \RobotRecovery{50} and \HumanRecovery{300}, but remains lower at
71.2. This controlled comparison shows that the proposed pathway uses human
recovery data more effectively than direct human recovery co-training. At the
same time, replacing too much robot recovery is detrimental: the setting with
\RobotRecovery{20} and \HumanRecovery{600} stays below
\RobotRecovery{50} and \HumanRecovery{300}. Panel (a) therefore identifies the
useful tradeoff, where human recovery improves coverage but still relies on a
nontrivial robot recovery anchor.

\textbf{Additional human recovery helps under low robot grounding, but does not replace it.}
Figure~\ref{fig:budget}(b) shows that with \RobotRecovery{20} fixed, adding
human recovery steadily improves \textsc{EgoRecovery} from 52.5 at
\HumanRecovery{100} to 75.0 at \HumanRecovery{600}, while the direct human
recovery mix improves more slowly from 45.0 to 62.5. This panel shows that
human recovery remains useful even when robot recovery grounding is limited.
However, the remaining gap to the stronger cost-matched setting with
\RobotRecovery{50} and \HumanRecovery{300} shows that more human recovery does
not fully compensate for too little robot recovery. Together, panels (a) and
(b) support the same conclusion: human recovery is most effective when it
augments, rather than replaces, robot recovery grounding.

\textbf{The throughput audit supports the data efficiency claim directly.}
Across the four tasks, robot success and robot recovery are collected at
roughly 50 episodes per operator hour, while human success and staged human
recovery are collected at roughly 500 episodes per operator hour. The exact
rates vary across tasks because staging and reset difficulty differ. On
average, however, human recovery is collected at approximately 10$\times$ the
robot recovery throughput. Together with Figure~\ref{fig:scaling}, this audit
supports the scaling claim by showing that the recovery data source with the
largest throughput is also the one that yields the clearest gains in Recovery
SR when transferred through \textsc{EgoRecovery}. Table~\ref{tab:throughput_audit}
reports the task-wise audit values.

\subsection{Ablation Study}
\label{sec:exp:ablation}

We next test whether the gain depends on the intended mechanism rather than on
the presence of a shared co-training backbone alone. Table~\ref{tab:ablation}
reports a compact closed-loop ablation using the same averaged Initial SR and
Recovery SR metrics as Figure~\ref{fig:main_results}. The direct human recovery
mix is included as the matched-data reference from the main comparison.
Appendix~\ref{appx:additional_ablation} complements this table with mechanism
diagnostics for the DCT target, the learned intent bottleneck, and the recovery
gate.

\begin{table}[t]
\vspace{-0.25em}
\centering
\begin{minipage}[t]{0.58\textwidth}
\vspace{0pt}
\centering
\caption{\textbf{Collection efficiency audit.} Accepted demonstrations per
operator hour under the data collection protocol.}
\label{tab:throughput_audit}
\vspace{\baselineskip}
\begingroup
\scriptsize
\setlength{\tabcolsep}{3.5pt}
\renewcommand{\arraystretch}{1.35}
\resizebox{\linewidth}{!}{%
\begin{tabular}{@{}lccccc@{}}
\toprule
Task
& \makecell{Robot\\success/hr}
& \makecell{Robot\\recovery/hr}
& \makecell{Human\\success/hr}
& \makecell{Human\\recovery/hr}
& \makecell{Human to Robot\\recovery} \\
\midrule
Cup brush insertion  & 58 & 48 & 483 & 522 & 10.9$\times$ \\
Table sweep          & 68 & 51 & 514 & 523 & 10.3$\times$ \\
Round disk placement & 62 & 45 & 522 & 511 & 11.4$\times$ \\
Cube stack           & 60 & 52 & 498 & 510 & 9.8$\times$ \\
\midrule
Average              & 62.0 & 49.0 & 504.3 & 516.5 & 10.5$\times$ \\
\bottomrule
\end{tabular}
}
\endgroup
\end{minipage}
\hfill
\begin{minipage}[t]{0.38\textwidth}
\vspace{0pt}
\centering
\caption{\textbf{Closed loop mechanism ablation.} Average Initial SR and
Recovery SR for the full method and controlled variants.}
\label{tab:ablation}
\begingroup
\scriptsize
\setlength{\tabcolsep}{4pt}
\renewcommand{\arraystretch}{1.12}
\resizebox{\linewidth}{!}{%
\begin{tabular}{@{}lcc@{}}
\toprule
Configuration & \makecell{Avg. Initial\\SR} & \makecell{Avg. Recovery\\SR} \\
\midrule
\textbf{Full EgoRecovery} & \textbf{80.0} & \textbf{85.0}  \\
Direct human recovery mix & 71.2 & 71.2 \\
without intent loss & 78.7 & 65.0 \\
without corrective segment mask & 75.0 & 55.0 \\
without intent modulation & 76.3 & 65.0 \\
$c_t$ zero at test time & 80.0 & 65.0 \\
always on modulation & 70.0 & 70.0 \\
\bottomrule
\end{tabular}
}
\endgroup
\end{minipage}
\vspace{-0.75em}
\end{table}

\textbf{Corrective intent supervision and recovery-specific masking both
matter.}
The full method matches the main result in Figure~\ref{fig:main_results} with 80.0 average
Initial SR and 85.0 average Recovery SR. Removing the intent loss reduces
average Recovery SR to 65.0, and removing the corrective segment mask degrades
it further to 55.0. This pattern is consistent with the method design. The
corrective intent bottleneck requires both the low-frequency DCT magnitude
target in Eq.~\ref{eq:intent_target} and a recovery-specific supervision region
that separates corrective motion from nominal motion.

\textbf{The learned intent must affect the action path.}
The direct human recovery mix reaches 71.2 average Recovery SR under the same
data composition as the full method. Thus, the presence of human recovery data
does not by itself explain the 85.0 result. When intent modulation is removed,
average Recovery SR drops to 65.0 even though intent can still be predicted as
an auxiliary output.
The $c_t$ zero intervention produces the same 20 percentage point drop from the full model.
This supports the interpretation that corrective intent must modulate the robot
action decoder to affect recovery, rather than serving only as a diagnostic
auxiliary representation.

\textbf{The Recovery Gate Head protects nominal execution.}
The always on modulation variant keeps recovery competitive but causes the
largest drop in average Initial SR, from 80.0 to 70.0. This is the expected
signature of the Recovery Gate Head. When recovery information is forced to
modulate all states, nominal execution becomes less stable.

\section{Conclusion}
\label{sec:conclusion}

We presented \textsc{EgoRecovery}, a framework for recovery learning that uses
egocentric human recovery as scalable supervision over failure states without treating
human motion as robot action. The method transfers only a compact corrective
intent, grounds it with limited robot recovery data, and activates it through a
learned recovery gate at deployment. Across four real robot tasks, this design
improves success from failure starts over robot-only recovery, direct
co-training with human recovery data, and ungated intent variants while preserving nominal execution.

\textbf{Limitations.}
These findings are limited to related off-nominal states within the same
task-level recovery family in tabletop manipulation. Human recovery reduces the
robot recovery data burden, but does not replace robot recovery grounding for
embodiment-specific contact handling, regrasping, or qualitatively new recovery
skills.

\clearpage
\bibliographystyle{plainnat}

\clearpage

\appendix

\section{Data Sources and Budgets}
\label{appx:data_budget}

All episode counts in the experiments are per task and source specific. They
indicate how many episodes of a given source are used when that source appears
in an experimental condition, rather than a single pooled training mixture or a
complete inventory of all collected data. The full \textsc{EgoRecovery} setting
uses Robot success~(50), Robot recovery~(50), and Human recovery~(300). Human
success~(300) is used only as the nominal human control in the main comparison.
The human data scaling study increases the human budget to 800 while keeping
the robot anchor fixed at Robot success~(50) and Robot recovery~(50). The
budget comparison additionally uses Robot recovery~(80) as a robot only
reference and Robot recovery~(20) as the low robot grounding setting.

\section{Closed-Loop Evaluation Protocol}
\label{appx:eval_protocol}

Each rollout is run for at most 5 minutes. A rollout that does not complete the
task within this window is counted as a failure. Policies may retry task actions
such as reinserting the brush, sweeping residual objects, or restacking a fallen
cube, provided that the robot remains engaged with the same task instance. A
rollout is terminated if the policy executes a reset action that returns the
robot to the initial pose or reinitializes the task, so recovery cannot be
achieved by erasing the failed attempt and restarting from scratch. For each
method, we use the same fixed set of 20 normal initial states and 20 common off
nominal recovery starts per task. In addition to success rates, we report the
corresponding success counts. Wilson 95\% binomial confidence intervals are
computed from the pooled success counts over the 80 rollouts that form each
four task aggregate.

\section{Task Success Criteria}
\label{appx:success_criteria}

All task outcomes are scored by a human evaluator from the synchronized rollout
videos. Scoring uses only the rollout videos and the rules below, without
recovery labels or corrective intent targets. A rollout is counted as successful
only if the task specific goal state is reached before the timeout and remains
visually stable for at least 3 s after the final corrective action. Multiple
attempts are allowed within the same rollout unless a task specific terminal
failure occurs or the policy executes a reset action as defined in
Appendix~\ref{appx:eval_protocol}.

\textbf{Cup brush insertion.} The brush is successful if the brush tip and shaft
are visibly inserted into the tube, the tube remains upright, and the brush does
not pop out or tilt out of the tube during the 3 s stability window.
Multiple insertion attempts are allowed before timeout. The rollout fails if the
tube is knocked over, the brush is dropped outside the reachable workspace, or
the robot resets the task.

\textbf{Table sweep.} The task is successful if all predefined trash objects are
inside the dustpan before pouring and are then transferred into the bin. Multiple
sweeping passes are allowed before pouring, but only one pouring action from the
dustpan to the bin is permitted. The rollout fails if any predefined trash
object remains on the table after pouring, if trash is spilled outside the bin
during pouring, or if the robot resets the task.

\textbf{Round disk placement.} The disk is successful if it is seated flush
inside the target groove, with no visible part of the disk protruding above the
rim or resting on the rim during the 3 s stability window. Multiple
placement attempts are allowed before timeout. The rollout fails if the disk is
dropped outside the reachable workspace, the target fixture is displaced, or the
robot resets the task.

\textbf{Cube stacking.} The task is successful if the yellow cube is placed on
top of the green cube and the stack remains upright for the 3 s stability
window. If the yellow cube falls before the final stable state, the robot may
retry stacking within the same rollout. The rollout fails if the green cube is
displaced from its original position, either cube leaves the reachable
workspace, or the robot resets the task.

\section{Common Off-Nominal Starts}
\label{appx:failure_regions}

Recovery SR evaluates policies from common off-nominal starts rather than from
the normal task initial state. These starts define the common recovery
distribution used for closed-loop evaluation. Robot and human recovery training
episodes are collected within the same task-level failure families, and the
evaluation starts use varied failed configurations from those same families
rather than reusing the exact robot recovery training starts. The benchmark
therefore tests transfer within related recovery behavior rather than unrelated
skills.
This section specifies the failed task state and
the robot arm state at the beginning of recovery evaluation. The recovery
rollout uses only robot observations. No external failure detector, human video,
or recovery label is provided at test time.

\textbf{Cup brush insertion.} The failed task state is that the brush has not
been inserted into the tube after an insertion attempt. Recovery starts with the
left and right arms holding the brush and tube, respectively, in a post-attempt
state where insertion has failed.

\textbf{Table sweep.} The failed task state is that the left-arm sweep has not
covered all trash objects, so trash remains on the table before the dustpan is
poured. Recovery starts with the left arm holding the broom and the right arm
holding the dustpan, after a partial sweep into the dustpan but before pouring.

\textbf{Round disk placement.} The failed task state is that the round disk has
not been inserted into the groove because it is misaligned with the target
groove. Recovery starts with the right arm holding the disk handle in a
post-placement-attempt state where the disk is near the groove but not seated.

\textbf{Cube stacking.} The failed task state is that the yellow cube has not
been stacked on the green cube after the stacking attempt. Recovery starts with
the robot arm hovering above the green cube after the failed placement attempt,
while the yellow cube remains grasped or within the reachable correction region.

\section{Data Collection and Annotation}
\label{appx:human_collection}

Together, this protocol defines the data collection and labeling process used
for the throughput audit in Table~\ref{tab:throughput_audit} and for the
recovery labels used in Section~\ref{sec:data}.

\paragraph{Recording setup.}
Human recordings use synchronized RGB-D streams at 25\,Hz. A head mounted
egocentric camera records the task scene and the demonstrator's first person
view. Two wrist mounted RGB-D cameras record close range hand object interaction
for the left and right hands. The current implementation uses Orbbec RGB-D
cameras, storing RGB frames as compressed images, depth frames as millimeter
scale uint16 maps, and intrinsics and timestamps for each episode.

\begin{figure}[H]
\centering
\includegraphics[width=0.92\textwidth]{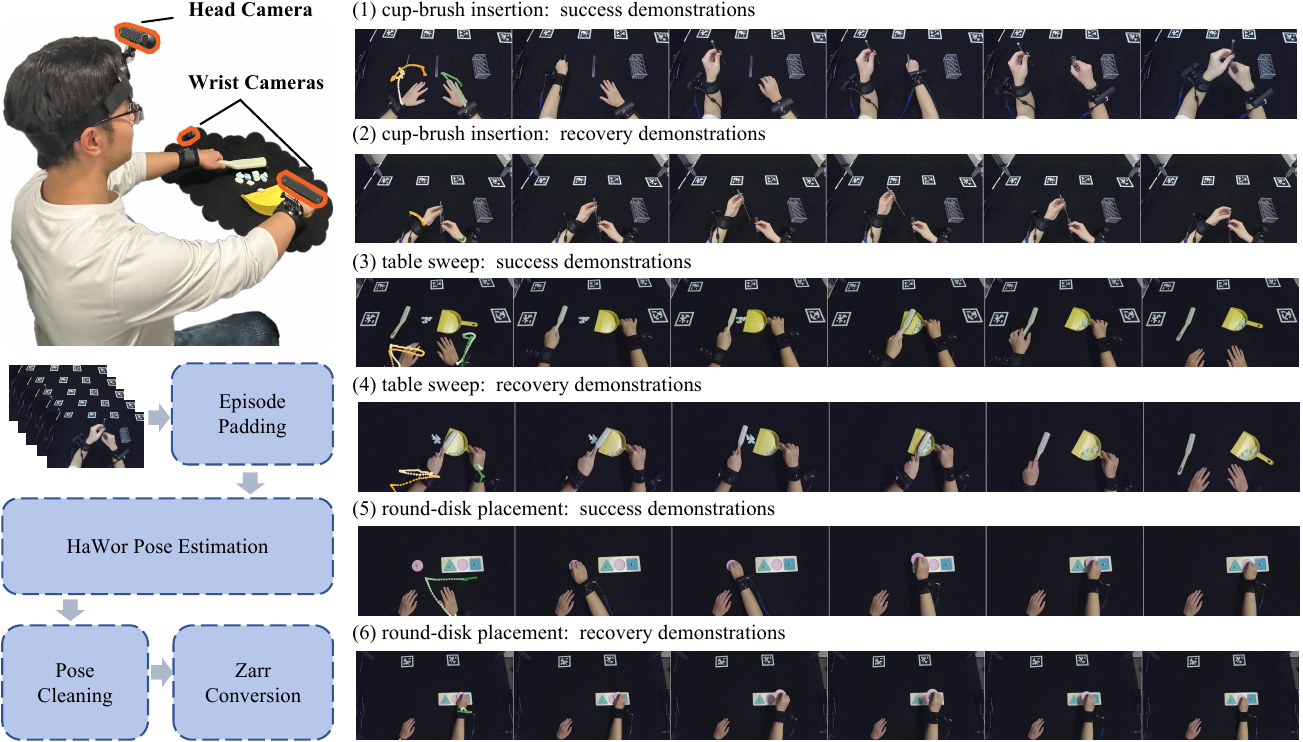}
\vspace{-0.35em}
\caption{\textbf{Egocentric human recovery collection setup.} Head mounted
first person recording is used to collect human recovery demonstrations from
staged off nominal starts, with wrist mounted streams recording close range
hand object interaction. These videos provide failure state and corrective
intent supervision during training only; policies are evaluated using robot
observations only.}
\label{fig:hand_setting}
\vspace{-0.45em}
\end{figure}

\begin{table}[H]
\centering
\caption{Human recording streams used for egocentric recovery collection. The
head mounted stream captures scene context and task progress, while the wrist
mounted streams capture close range hand object interaction. All streams store
RGB, intrinsics, and timestamps for synchronization.}
\label{tab:human_sensors}
\resizebox{\textwidth}{!}{%
\begin{tabular}{llll}
\toprule
Stream & Sensor position & Main use & Stored data \\
\midrule
Ego RGB & Head mounted & Scene context and task progress & RGB,  intrinsics, timestamps \\
Left wrist RGB & Left wrist & Hand object contact & RGB,  intrinsics, timestamps \\
Right wrist RGB & Right wrist & Hand object contact & RGB, intrinsics, timestamps \\
\bottomrule
\end{tabular}}
\end{table}

\paragraph{Common off nominal collection protocol.}
When used as a nominal human control, human success demonstrations start from
ordinary task initial states and run to task completion. Human and robot
recovery episodes use the task level failure families in
Appendix~\ref{appx:failure_regions}, with varied failed configurations inside
each family. Human recovery clips contain only the corrective segment after a
staged failure state. For the throughput accounting in
Table~\ref{tab:throughput_audit}, setup time, resets, synchronization, and
discarded attempts are counted for both human and robot collection.

\paragraph{Episode filtering.}
An episode is accepted if it begins from a valid failure state, contains a
complete corrective maneuver, and has synchronized observations throughout the
active phase. Human episodes are additionally filtered for reliable bilateral
hand tracking and the absence of severe occlusion or sensor dropout. Robot
episodes are filtered for synchronized robot state and action logs, no major
teleoperation interruption, and no severe sensor dropout. Discarded episodes are
not used for training but remain part of the operator hour accounting.

\paragraph{Recovery segment annotation.}
\label{appx:recovery_annotation}
Each accepted recovery clip starts at a failure state, so the annotator marks
only one boundary, $t_{\mathrm{rec}}$, where corrective motion ends and ordinary
task execution resumes. We obtain this boundary in two steps. First, we smooth
the hand or end effector motion and compute a framewise motion energy curve,
which gives an automatic candidate for $t_{\mathrm{rec}}$. The annotator then
checks the synchronized video together with the curve and either confirms the
candidate or moves it to the frame where corrective motion has returned to
ordinary execution. The same pass records a quality flag and whether the episode
should be discarded.

For accepted episodes, $t_{\mathrm{rec}}$ is converted into the labels
summarized in Table~\ref{tab:annotation_schema}. The recovery state label $s_t$
marks the recovery phase for the Recovery Gate Head. The intent mask $m_t$ is a
stricter training mask whose validity checks are detailed in
Appendix~\ref{appx:intent_target}. These labels are used only during training
and are not available at deployment.

\begin{table}[H]
\centering
\caption{Recovery annotation schema used to convert each accepted recovery
episode into training labels. The episode boundary defines the recovery phase,
while the derived frame labels supervise the Recovery Gate Head and the
corrective intent loss.}
\label{tab:annotation_schema}
\begin{tabular}{lll}
\toprule
Field & Level & Use \\
\midrule
$t_{\mathrm{rec}}$ & Episode & Boundary between recovery and normal execution \\
\texttt{quality}, \texttt{discard} & Episode & Filtering and auditing \\
$s_t$ & Frame & Target for the Recovery Gate Head \\
$m_t$ & Frame or segment & Mask for corrective intent supervision \\
\bottomrule
\end{tabular}
\end{table}

\section{Motion Reconstruction and Data Preprocessing}
\label{appx:preprocessing}

\paragraph{Hand trajectory reconstruction.}
We estimate bilateral 3D hand motion from the egocentric video using
HaWoR~\citep{hawor2025}. From the reconstructed MANO hands we extract the
wrist joint position and root orientation. The orientation is converted to
roll-pitch-yaw angles, giving a 6-DoF wrist pose $[x,y,z,r_x,r_y,r_z]$ in
meters and radians. Together with the gripper aperture proxy below, each hand is
represented as $[x,y,z,r_x,r_y,r_z,\bar{g}]$, giving a 14D bilateral human state
stream. This reconstructed pose stream is used as the human embodiment
trajectory for the auxiliary human decoder and for corrective intent target
construction. It is not used as robot frame action supervision.

We compute a gripper-aperture proxy from fingertip distance:
\begin{equation}
  g_t = \bigl\|p_{\mathrm{thumb}}^t - p_{\mathrm{index}}^t\bigr\|_2,
  \label{eq:gripper}
\end{equation}
where $p_{\mathrm{thumb}}^t$ and $p_{\mathrm{index}}^t$ are the MANO fingertip
positions. This distance is linearly mapped to $\bar{g}_t\in[0,1]$ over
$[0.03\,\mathrm{m},0.08\,\mathrm{m}]$, with clipping.

\paragraph{Filtering and resampling.}
Frames with wrist position jumps above 30\,mm or Euler-angle jumps above
$20^\circ$ are replaced by interpolation from neighboring valid frames. We then
apply median filtering and Savitzky--Golay smoothing; rotation channels are
unwrapped before smoothing and re-wrapped afterwards. Continuous state vectors
are exported to the 25\,Hz trajectory format used by the training zarrs through
linear interpolation; video frames are selected by nearest neighbor. Segments
that remain unreliable after filtering are rejected or masked out before intent
supervision.

\section{Corrective Intent Target}
\label{appx:intent_target}

\paragraph{Target construction.}
The intent loss uses the target only where the intent mask $m_t$ is true. We
first define target construction and then define the validity mask. Following
Eq.~\ref{eq:intent_target}, we compute the DCT magnitude target from the scaled
relative future motion of each active end effector. This appendix specifies the
concrete windowing and masking choices used in the experiments. Raw trajectories
are stored at 25\,Hz, but the intent target is computed on a phase resampled
grid rather than on a fixed duration window. Each episode is linearly resampled
to $P=200$ phase points. For an anchor frame $t$, we map the frame to its phase
index $p_t$ and take a future window of $L=16$ phase points, which covers 8\% of
that episode. Thus the window duration varies with episode length. For example,
it is about 0.4\,s for a 5\,s human recovery clip.

The main experiments use $K=4$ DCT coefficients and one active end effector,
resulting in a 4D target. The active effector is the task designated active arm,
while the other arm holds a static grip pose when applicable. The same
construction supports two active effectors, which would give an 8D target, but
that two effector variant is not used in the main results.

\paragraph{Masking and normalization.}
The effective intent mask is the conjunction of two masks. The first mask,
\texttt{gt\_intent\_valid}, requires the frame to be unpadded, the future window
to fit inside the phase resampled episode, and the peak to peak $\ell_2$
displacement of the window to be at least 0.01\,m, which removes near stationary
segments. The second mask, \texttt{recovery\_intent\_valid}, requires the anchor
frame to lie inside the annotated recovery phase and the full future window to
remain inside that phase. Success episodes set both masks to false. The loss
uses their conjunction, so intent supervision is applied only to valid recovery
windows.

Invalid segments are stored as masked targets during preprocessing and do not
contribute to the intent loss or normalization statistics. Consistent with the
main text target in Eq.~\ref{eq:intent_target}, embodiment scale calibration is
applied before the DCT through $s_d$. Normalization statistics are computed only
from valid recovery windows, so padded frames, invalid windows, and success
episode targets are excluded.

\section{Training Details}
\label{appx:training}

The training implementation follows the heterogeneous policy training setup
used in EgoMimic style co training, with a shared visual and Transformer trunk
and embodiment specific observation adapters and action decoders. The human
decoder is trained only with reconstructed human hand trajectories, while the
robot decoder is trained only with robot teleoperation actions. Thus human
action labels help train the shared representation and human decoder but are not
used as robot action supervision.

Training uses Robot and Human dataloaders whose selected episode pools follow
the budget of each experimental condition. For the main comparison, this means
\RobotSuccess{50}, \RobotRecovery{50}, and \HumanRecovery{300}; Human success
episodes are used only in the corresponding human success control. Frames are
sampled uniformly within each selected pool, without an additional class
balanced or phase balanced sampler. The direct mix baseline and mechanism
ablations use the same sampling and optimization settings unless the ablation
explicitly removes a data source or disables one model component.

Robot actions are 14D absolute joint position targets, with six arm joints and
one gripper value per arm. Human actions are 14D absolute hand state targets,
with $x,y,z,r_x,r_y,r_z$ and a gripper proxy per hand. Both streams are stored
at 25\,Hz and are used to form 100 step action chunks for the corresponding
decoder. We use the objective in Eq.~\ref{eq:training_objective}; the appendix
reports the implementation constants below rather than repeating the gated
modulation equations from the main text.

\begin{table}[H]
\centering
\caption{Compact training details for the main \textsc{EgoRecovery} runs.}
\label{tab:training_hparams}
\begingroup
\small
\setlength{\tabcolsep}{6pt}
\renewcommand{\arraystretch}{1.10}
\begin{tabularx}{0.90\textwidth}{@{}>{\bfseries\raggedright\arraybackslash}p{0.24\textwidth}X@{}}
\toprule
\multicolumn{2}{@{}l}{\textit{Model}} \\
\addlinespace[0.15em]
Policy & HPT style co training \\
Backbone & 16 blocks, $d=256$, 8 heads, drop path 0.1 \\
Decoder & 8 layers, 100 step horizon, 14D output \\
Heads & 4D intent, 1D gate, gated FiLM \\
\midrule
\multicolumn{2}{@{}l}{\textit{Optimization}} \\
\addlinespace[0.15em]
Batch & 24 per GPU, 192 global \\
Optimizer & AdamW, lr $10^{-4}$, wd $10^{-4}$ \\
Scheduler & Cosine, $T_{\max}=200$, min lr $2\times10^{-6}$ \\
Loss & $\lambda_c=0.05$, $\lambda_g=0.05$, $\lambda_n=0.01$, $\beta=0.05$ \\
\midrule
\multicolumn{2}{@{}l}{\textit{Input processing}} \\
\addlinespace[0.15em]
Augmentation & Color jitter, ImageNet norm \\
\bottomrule
\end{tabularx}
\endgroup
\end{table}

All main models are trained for 20{,}000 optimization steps with BF16 mixed
precision on 8 NVIDIA L20 GPUs. Checkpoints are saved every 10{,}000 steps, and
reported rollouts use the final 20{,}000 step checkpoint unless otherwise noted.
The implementation uses dynamic median absolute deviation gradient clipping
after a 100 step warmup.

For compute accounting, the primary paper-reported runs were trained on a single
8 GPU node with 48 GB NVIDIA L20 devices. A typical main co-training run took
about 10 hours, or roughly 80 L20 GPU hours. The training runs reported in the
paper required approximately 2000 L20 GPU hours in total, with exploratory,
failed, or superseded runs during method development adding on the order of a
few hundred additional L20 GPU hours. Each reported model was trained with a
single seed. Peak memory was not logged per run, but all reported runs fit within
the 48 GB per GPU memory budget under BF16 mixed precision.

\section{Full Closed Loop Statistics for Main Results}
\label{appx:additional_results}

Following the closed loop protocol in Appendix~\ref{appx:eval_protocol},
Table~\ref{tab:full_main_results} reports the pooled success counts and Wilson
confidence intervals behind the main comparison. The task wise percentages are
shown in Figure~\ref{fig:main_results}; this table provides the count level
audit. Table~\ref{tab:full_scaling_results} provides the same audit for the
human data scaling study in Figure~\ref{fig:scaling}, while
Tables~\ref{tab:full_budget_cost_matched} and
\ref{tab:full_budget_low_robot} report the two regimes in
Figure~\ref{fig:budget}. All statistics in this section pool the four tasks
used for the corresponding main text average. An entry such as
$80.0\,[70.0,87.3]$ reports the success rate in percent followed by the Wilson
95\% binomial confidence interval.

These tables are intended as an audit of the main figures rather than as new
experimental settings. The count-level view supports the same conclusion as the
main text. \textsc{EgoRecovery} improves recovery most reliably when human
recovery data are combined with a nontrivial robot recovery anchor, while direct
human recovery co training and excessive replacement of robot recovery are
weaker.

\begin{table}[H]
\centering
\caption{Aggregate statistics for the main closed loop comparison.
All methods are evaluated on the same fixed set of 20 initial states and 20
recovery starts per task. Counts aggregate the four tasks, giving 80 rollouts
per start state regime. Intervals are Wilson 95\% binomial confidence
intervals. Bold values mark the best aggregate result.}
\label{tab:full_main_results}
\begingroup
\small
\setlength{\tabcolsep}{4pt}
\renewcommand{\arraystretch}{1.18}
\resizebox{0.98\textwidth}{!}{%
\begin{tabular}{@{}lcccccccc@{}}
\toprule
Method
& \makecell{Robot\\success}
& \makecell{Robot\\recovery}
& \makecell{Human\\success}
& \makecell{Human\\recovery}
& \makecell{Initial\\successes}
& \makecell{Initial SR\\(\%, Wilson CI)}
& \makecell{Recovery\\successes}
& \makecell{Recovery SR\\(\%, Wilson CI)} \\
\midrule
\makecell[l]{Robot success only}
& 50 & -- & -- & --
& 30/80 & 37.5 [27.7, 48.5] & 0/80 & 0.0 [0.0, 4.6] \\
\makecell[l]{$R_{\mathrm{suc}}(50) + H_{\mathrm{suc}}(300)$}
& 50 & -- & 300 & --
& 36/80 & 45.0 [34.6, 55.9] & 0/80 & 0.0 [0.0, 4.6] \\
\makecell[l]{$R_{\mathrm{suc}}(50) + H_{\mathrm{rec}}(300)$}
& 50 & -- & -- & 300
& 34/80 & 42.5 [32.3, 53.4] & 7/80 & 8.8 [4.3, 17.0] \\
\makecell[l]{Robot recovery}
& 50 & 50 & -- & --
& 45/80 & 56.3 [45.3, 66.6] & 42/80 & 52.5 [41.7, 63.1] \\
\makecell[l]{Human success mix}
& 50 & 50 & 300 & --
& 51/80 & 63.8 [52.8, 73.4] & 52/80 & 65.0 [54.1, 74.5] \\
\makecell[l]{Direct human recovery mix}
& 50 & 50 & -- & 300
& 57/80 & 71.2 [60.5, 80.0] & 57/80 & 71.2 [60.5, 80.0] \\
\textbf{\textsc{EgoRecovery}}
& \textbf{50} & \textbf{50} & \textbf{--} & \textbf{300}
& \textbf{64/80} & \textbf{80.0 [70.0, 87.3]}
& \textbf{68/80} & \textbf{85.0 [75.6, 91.2]} \\
\bottomrule
\end{tabular}
}
\endgroup
\end{table}

\begin{table}[H]
\centering
\caption{Pooled statistics for the human data scaling study in
Figure~\ref{fig:scaling}. All rows use the fixed robot anchor
\RobotSuccess{50} and \RobotRecovery{50}. Counts aggregate four tasks with 80
rollouts per start state regime. Bold values mark the better matched condition
at each human data budget.}
\label{tab:full_scaling_results}
\begingroup
\small
\setlength{\tabcolsep}{3.5pt}
\renewcommand{\arraystretch}{1.12}
\resizebox{0.98\textwidth}{!}{%
\begin{tabular}{@{}lcccccccc@{}}
\toprule
Method
& \makecell{Robot\\success}
& \makecell{Robot\\recovery}
& \makecell{Human\\success}
& \makecell{Human\\recovery}
& \makecell{Initial\\successes}
& \makecell{Initial SR\\(\%, Wilson CI)}
& \makecell{Recovery\\successes}
& \makecell{Recovery SR\\(\%, Wilson CI)} \\
\midrule
Human success co training & 50 & 50 & 200 & -- & 44/80 & 55.0 [44.1, 65.4] & 47/80 & 58.8 [47.8, 68.9] \\
\textsc{EgoRecovery} & 50 & 50 & -- & 200 & \textbf{59/80} & \textbf{73.8 [63.2, 82.1]} & \textbf{60/80} & \textbf{75.0 [64.5, 83.2]} \\
\addlinespace[0.12em]
Human success co training & 50 & 50 & 300 & -- & 51/80 & 63.8 [52.8, 73.4] & 52/80 & 65.0 [54.1, 74.5] \\
\textsc{EgoRecovery} & 50 & 50 & -- & 300 & \textbf{64/80} & \textbf{80.0 [70.0, 87.3]} & \textbf{68/80} & \textbf{85.0 [75.6, 91.2]} \\
\addlinespace[0.12em]
Human success co training & 50 & 50 & 500 & -- & 56/80 & 70.0 [59.2, 78.9] & 53/80 & 66.2 [55.4, 75.7] \\
\textsc{EgoRecovery} & 50 & 50 & -- & 500 & \textbf{67/80} & \textbf{83.8 [74.2, 90.3]} & \textbf{68/80} & \textbf{85.0 [75.6, 91.2]} \\
\addlinespace[0.12em]
Human success co training & 50 & 50 & 800 & -- & 59/80 & 73.8 [63.2, 82.1] & 55/80 & 68.8 [57.9, 77.8] \\
\textsc{EgoRecovery} & 50 & 50 & -- & 800 & \textbf{70/80} & \textbf{87.5 [78.5, 93.1]} & \textbf{72/80} & \textbf{90.0 [81.5, 94.8]} \\
\bottomrule
\end{tabular}
}
\endgroup
\end{table}

\begin{table}[H]
\centering
\caption{Pooled recovery statistics for the cost matched substitution regime in
Figure~\ref{fig:budget}(a). The robot equivalent recovery budget is fixed at
$\Beq=80$. Counts aggregate four tasks with 80 recovery rollouts per row. Bold
values mark the higher result for each matched recovery budget.}
\label{tab:full_budget_cost_matched}
\begingroup
\small
\setlength{\tabcolsep}{3.5pt}
\renewcommand{\arraystretch}{1.10}
\resizebox{0.98\textwidth}{!}{%
\begin{tabular}{@{}lccccccc@{}}
\toprule
Method
& \makecell{Robot\\success}
& \makecell{Robot\\recovery}
& \makecell{Human\\success}
& \makecell{Human\\recovery}
& $\Beq$
& \makecell{Recovery\\successes}
& \makecell{Recovery SR\\(\%, Wilson CI)} \\
\midrule
Robot only reference & 50 & 80 & -- & -- & 80 & 48/80 & 60.0 [49.0, 70.0] \\
Direct human recovery mix & 50 & 70 & -- & 100 & 80 & 49/80 & 61.2 [50.3, 71.2] \\
\textsc{EgoRecovery} & 50 & 70 & -- & 100 & 80 & \textbf{56/80} & \textbf{70.0 [59.2, 78.9]} \\
\addlinespace[0.12em]
Direct human recovery mix & 50 & 60 & -- & 200 & 80 & 52/80 & 65.0 [54.1, 74.5] \\
\textsc{EgoRecovery} & 50 & 60 & -- & 200 & 80 & \textbf{61/80} & \textbf{76.2 [65.9, 84.2]} \\
\addlinespace[0.12em]
Direct human recovery mix & 50 & 50 & -- & 300 & 80 & 57/80 & 71.2 [60.5, 80.0] \\
\textsc{EgoRecovery} & 50 & 50 & -- & 300 & 80 & \textbf{68/80} & \textbf{85.0 [75.6, 91.2]} \\
\addlinespace[0.12em]
Direct human recovery mix & 50 & 40 & -- & 400 & 80 & 53/80 & 66.2 [55.4, 75.7] \\
\textsc{EgoRecovery} & 50 & 40 & -- & 400 & 80 & \textbf{62/80} & \textbf{77.5 [67.2, 85.3]} \\
\addlinespace[0.12em]
Direct human recovery mix & 50 & 30 & -- & 500 & 80 & 50/80 & 62.5 [51.5, 72.3] \\
\textsc{EgoRecovery} & 50 & 30 & -- & 500 & 80 & \textbf{58/80} & \textbf{72.5 [61.9, 81.1]} \\
\addlinespace[0.12em]
Direct human recovery mix & 50 & 20 & -- & 600 & 80 & 50/80 & 62.5 [51.5, 72.3] \\
\textsc{EgoRecovery} & 50 & 20 & -- & 600 & 80 & \textbf{60/80} & \textbf{75.0 [64.5, 83.2]} \\
\bottomrule
\end{tabular}
}
\endgroup
\end{table}

\begin{table}[H]
\centering
\caption{Pooled recovery statistics for the low robot grounding regime in
Figure~\ref{fig:budget}(b). Robot recovery is fixed at \RobotRecovery{20}, and
human recovery is varied. Counts aggregate four tasks with 80 recovery rollouts
per row. Bold values mark the higher result for each human recovery budget.}
\label{tab:full_budget_low_robot}
\begingroup
\small
\setlength{\tabcolsep}{3.5pt}
\renewcommand{\arraystretch}{1.10}
\resizebox{0.98\textwidth}{!}{%
\begin{tabular}{@{}lccccccc@{}}
\toprule
Method
& \makecell{Robot\\success}
& \makecell{Robot\\recovery}
& \makecell{Human\\success}
& \makecell{Human\\recovery}
& $\Beq$
& \makecell{Recovery\\successes}
& \makecell{Recovery SR\\(\%, Wilson CI)} \\
\midrule
Robot only reference & 50 & 20 & -- & -- & 20 & 28/80 & 35.0 [25.5, 45.9] \\
Direct human recovery mix & 50 & 20 & -- & 100 & 30 & 36/80 & 45.0 [34.6, 55.9] \\
\textsc{EgoRecovery} & 50 & 20 & -- & 100 & 30 & \textbf{42/80} & \textbf{52.5 [41.7, 63.1]} \\
\addlinespace[0.12em]
Direct human recovery mix & 50 & 20 & -- & 200 & 40 & 42/80 & 52.5 [41.7, 63.1] \\
\textsc{EgoRecovery} & 50 & 20 & -- & 200 & 40 & \textbf{48/80} & \textbf{60.0 [49.0, 70.0]} \\
\addlinespace[0.12em]
Direct human recovery mix & 50 & 20 & -- & 300 & 50 & 46/80 & 57.5 [46.6, 67.7] \\
\textsc{EgoRecovery} & 50 & 20 & -- & 300 & 50 & \textbf{53/80} & \textbf{66.2 [55.4, 75.7]} \\
\addlinespace[0.12em]
Direct human recovery mix & 50 & 20 & -- & 400 & 60 & 48/80 & 60.0 [49.0, 70.0] \\
\textsc{EgoRecovery} & 50 & 20 & -- & 400 & 60 & \textbf{56/80} & \textbf{70.0 [59.2, 78.9]} \\
\addlinespace[0.12em]
Direct human recovery mix & 50 & 20 & -- & 500 & 70 & 50/80 & 62.5 [51.5, 72.3] \\
\textsc{EgoRecovery} & 50 & 20 & -- & 500 & 70 & \textbf{58/80} & \textbf{72.5 [61.9, 81.1]} \\
\addlinespace[0.12em]
Direct human recovery mix & 50 & 20 & -- & 600 & 80 & 50/80 & 62.5 [51.5, 72.3] \\
\textsc{EgoRecovery} & 50 & 20 & -- & 600 & 80 & \textbf{60/80} & \textbf{75.0 [64.5, 83.2]} \\
\bottomrule
\end{tabular}
}
\endgroup
\end{table}

\section{Additional Ablations and Mechanism Analysis}
\label{appx:additional_ablation}

The closed loop ablations in Table~\ref{tab:ablation} test whether the recovery
gated intent pathway is needed for policy performance. This section complements
those ablations with mechanism diagnostics that inspect whether the internal
signals behave as intended. The diagnostics follow the method design in
Sections~\ref{sec:factor} and~\ref{sec:align}: first, whether the DCT intent
target carries a comparable recovery signal across embodiments; second, whether
the trained Intent Head maps human and robot recovery observations to a shared
4D representation; and third, whether the Recovery Gate activates near the
annotated recovery boundary. We first test the target itself by comparing the
human and robot distributions of the same $K=4$ low-frequency DCT magnitude
features used as the intent target in Eq.~\ref{eq:intent_target}. We use the
two-sample Kolmogorov--Smirnov (KS) distance for each coefficient
$k\in\{0,1,2,3\}$; lower KS indicates stronger cross-embodiment distributional
overlap.

\begin{figure}[H]
\centering
\includegraphics[width=0.96\textwidth]{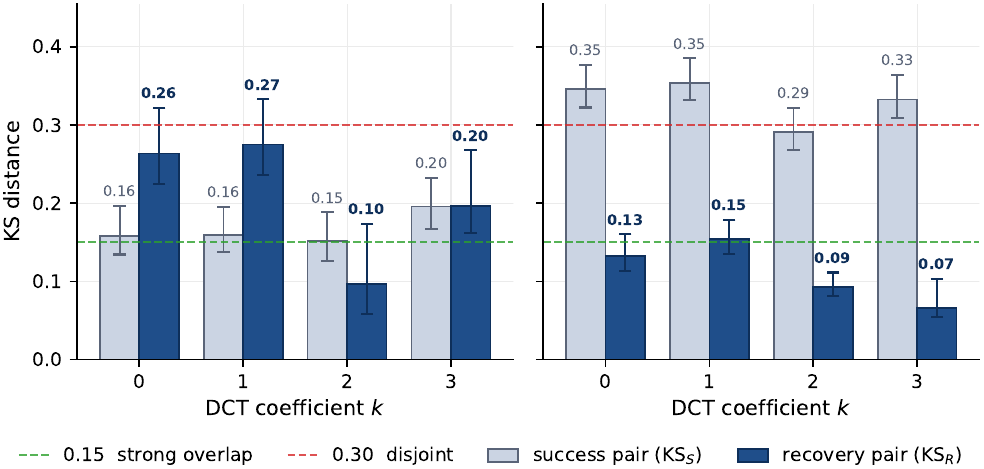}
\vspace{-0.35em}
\caption{\textbf{Cross-embodiment alignment of the $K=4$ DCT magnitude intent
target.} Bars report two-sample KS distances between robot and human DCT target
distributions for each coefficient. Light bars compare success trajectories and
dark bars compare recovery sub-trajectories; lower values indicate stronger
cross-embodiment overlap. Left and right panels show cup brush insertion and
round disk placement, respectively.}
\label{fig:ks_intent_alignment}
\vspace{-0.65em}
\end{figure}

The KS diagnostic shows that the recovery-window DCT target is useful but not
uniformly strong across tasks. Whiskers show 95\% bootstrap confidence intervals
($n_{\mathrm{boot}}=1{,}000$, seed 42); the green and red reference lines mark
0.15 strong overlap and 0.30 weak overlap references, respectively. Round disk placement is
close to the intended case: the full success trajectories remain embodiment
specific, while the recovery-window DCT magnitude features become substantially
more aligned ($\mathrm{KS}_{\mathrm{R}}=0.11$ vs.
$\mathrm{KS}_{\mathrm{S}}=0.33$). Cup brush insertion is weaker
($\mathrm{KS}_{\mathrm{R}}=0.21$, $\mathrm{KS}_{\mathrm{S}}=0.17$). Its recovery
often contains small left-arm translation and larger rotational adjustment, so
an xyz-only DCT magnitude feature can retain embodiment-specific micro-motion
differences. This motivates inspecting the learned bottleneck and Recovery Gate
Head rather than relying on the target distribution alone.

We next inspect whether the trained Intent Head keeps the intended
cross-embodiment structure in its learned bottleneck. For each frame, we compare
the pooled trunk feature that enters the Intent Head (256D) with the predicted
intent latent $c_t$ (4D). We keep recovery frames and nominal frames with valid
intent windows, excluding pure-success frames that do not receive intent
supervision. Distances are computed in the original feature spaces, not in the
2D visualization. Let $W_2^R$ denote the sliced Wasserstein-2 distance between
robot and human recovery features, and let $W_2^N$ denote the corresponding
distance for nominal features. The ratio
$\rho = W_2^R / W_2^N$ measures whether recovery states are more
cross-embodiment aligned than nominal states.

\begin{figure}[H]
\centering
\includegraphics[width=0.86\textwidth]{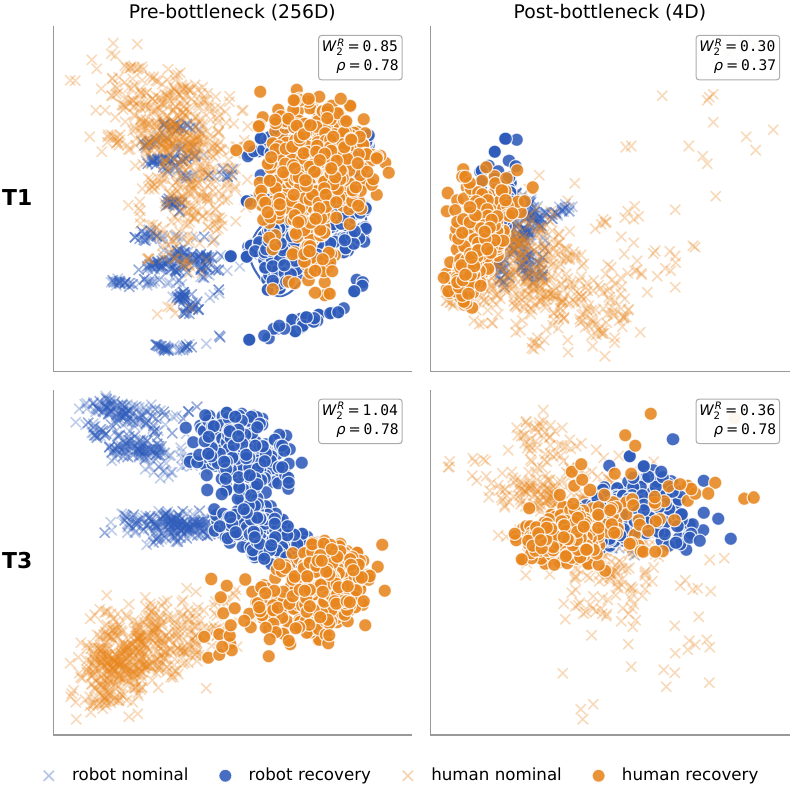}
\vspace{-0.35em}
\caption{\textbf{Learned alignment in the corrective-intent bottleneck.}
PCA-2 visualizations of the pre-bottleneck Intent Head input (left) and the
trained 4D intent latent $c_t$ (right) for cup brush insertion (T1) and round
disk placement (T3). Color denotes embodiment and marker shape denotes phase.
The boxed values are computed in the original feature space rather than the
PCA plane: $W_2^R$ is the sliced Wasserstein-2 distance between robot and human
recovery features, and $\rho=W_2^R/W_2^N$ compares recovery alignment with
nominal alignment.}
\label{fig:intent_latent_alignment}
\vspace{-0.65em}
\end{figure}

Figure~\ref{fig:intent_latent_alignment} supports the role assigned to
$c_t$ in the main method. The absolute robot-human recovery distance becomes
much smaller after the trained bottleneck on both tasks, while random JL
projection and PCA-4 provide little or moderate shrinkage after correcting for
dimension. This indicates that the alignment is learned from the
corrective-intent supervision rather than being a direct consequence of reducing
256 dimensions to 4. We use dimension-normalized sliced Wasserstein distances,
because raw sliced Wasserstein estimates have a $1/\sqrt{d}$ dependence. Under
this normalization, the trained bottleneck reduces robot-human recovery distance
by 22.5$\times$ on cup brush and 23.2$\times$ on round disk. The corresponding
random JL projection gives only 1.2$\times$/1.1$\times$, and PCA-4 gives
2.0$\times$/1.1$\times$. The ratio $\rho$ gives a complementary reading. In both
tasks, recovery features are more aligned across embodiments than nominal features
($\rho<1$), consistent with the data-collection goal of sampling related
failure states. The bottleneck further reduces this ratio on cup brush
($0.78\rightarrow0.37$), but leaves it nearly unchanged on round disk
($0.78\rightarrow0.78$). We therefore treat the absolute distance reduction as
the task-consistent evidence for the shared intent representation, and the
change in $\rho$ as a task-dependent effect. On round disk, PCA-4 is almost
identical to random projection, while the trained bottleneck still gives a
large reduction; this suggests that the recovery alignment is not captured by
the top-variance directions of the pre-bottleneck feature.

Finally, we examine the gate by aligning episodes to the human-labeled recovery
boundary $t_{\mathrm{recover}}$ and plotting the predicted recovery probability
$p_{\mathrm{rec}}$ for robot and human demonstrations. The phase-only baseline
uses only normalized frame position, so it measures how much of the recovery
label can be inferred from episode timing without visual input.

\begin{figure}[H]
\centering
\includegraphics[width=0.96\textwidth]{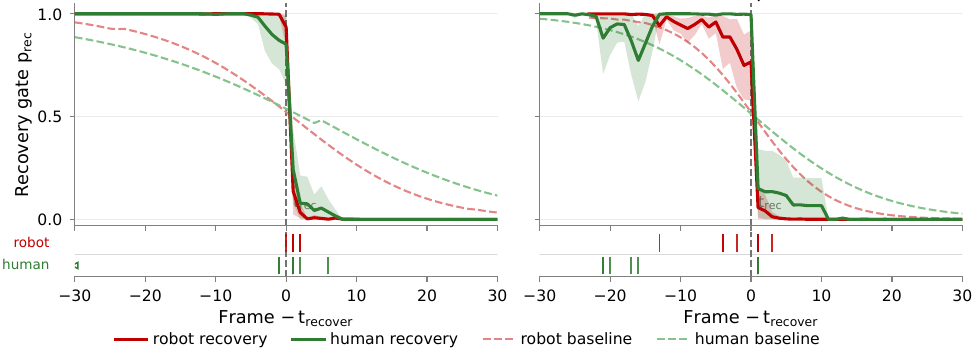}
\vspace{-0.35em}
\caption{\textbf{Recovery Gate Head aligned to annotated recovery boundaries.}
Predictions $p_{\mathrm{rec}}(\cdot)$ are aligned to
$t_{\mathrm{recover}}$ on 15 robot and 15 human recovery episodes per task.
Solid curves show per-domain means with 95\% bootstrap confidence intervals,
and dashed curves show a phase-only baseline. Left and right panels show cup
brush insertion and round disk placement. The gate drops near the annotated
transition for both embodiments, limiting intent modulation to recovery frames.}
\label{fig:gate_diagnostics}
\vspace{-0.65em}
\end{figure}

Figure~\ref{fig:gate_diagnostics} contains four pieces of evidence: the mean
gate output over aligned episodes, bootstrap uncertainty across episodes, a
timing-only reference, and per-episode threshold-crossing offsets. Together,
these diagnostics support the intended role of the gate. In both tasks,
$p_{\mathrm{rec}}$ is high before $t_{\mathrm{recover}}$ and drops near the
annotated transition, allowing corrective intent to affect recovery frames
without remaining active during ordinary execution. The phase-only baseline is
a logistic regression on normalized frame position, and bottom ticks mark the
first frame in each episode where $p_{\mathrm{rec}}<0.5$. Cup brush shows the
clearest visual gain over this timing-only reference (gate AUC 0.9998 vs.
phase-only AUC 0.980/0.969 for robot/human). All 30 robot and human episodes
cross the $0.5$ threshold within $\pm1$ frame of $t_{\mathrm{recover}}$,
indicating precise boundary alignment despite the weaker KS alignment in
Figure~\ref{fig:ks_intent_alignment}. Round disk is more conservative evidence:
the phase-only baseline is already near saturated (0.994/0.996) because
recovery timing is highly regular, and human episodes show a wider
crossing-offset spread (IQR $[-16.5,+1]$ frames), consistent with larger
hand-motion variability. This diagnostic therefore supports gate alignment,
while the closed loop ablation remains the causal test that the gate improves
deployed robot behavior.

\subsection{Data Collection and Rollout Visualization}
\label{appx:qualitative}

This section provides a visual overview of the data collection pipeline and the
closed-loop \textsc{EgoRecovery} inference results. Tasks 1--4 correspond to
cup-brush insertion, table sweep, round-disk placement, and cube stacking,
respectively.

\subsubsection{Data Collection}

\begin{figure}[H]
\centering
\includegraphics[width=\textwidth]{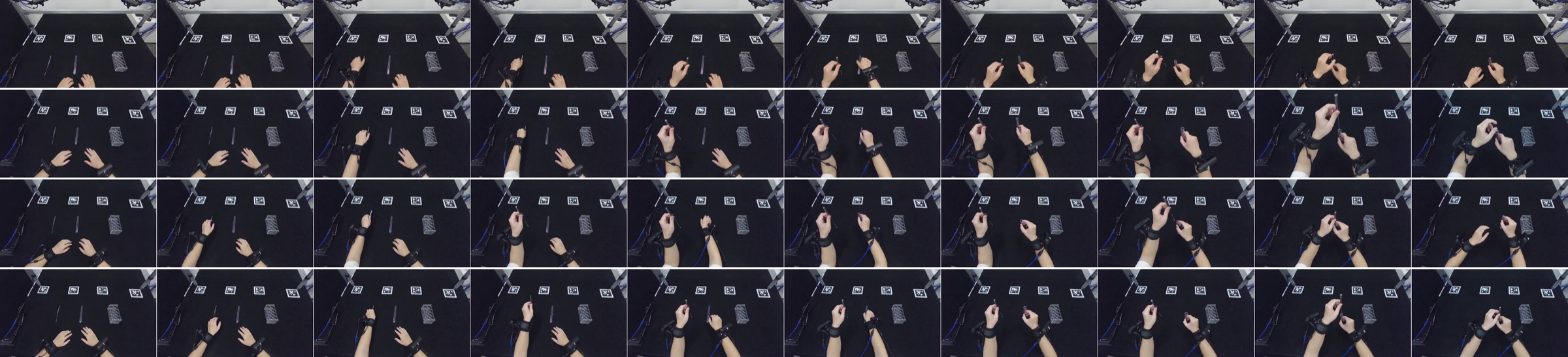}
\caption{Human success collection for cup-brush insertion (Task 1).}
\label{fig:collection_t1_hand_success}
\end{figure}

\begin{figure}[H]
\centering
\includegraphics[width=\textwidth]{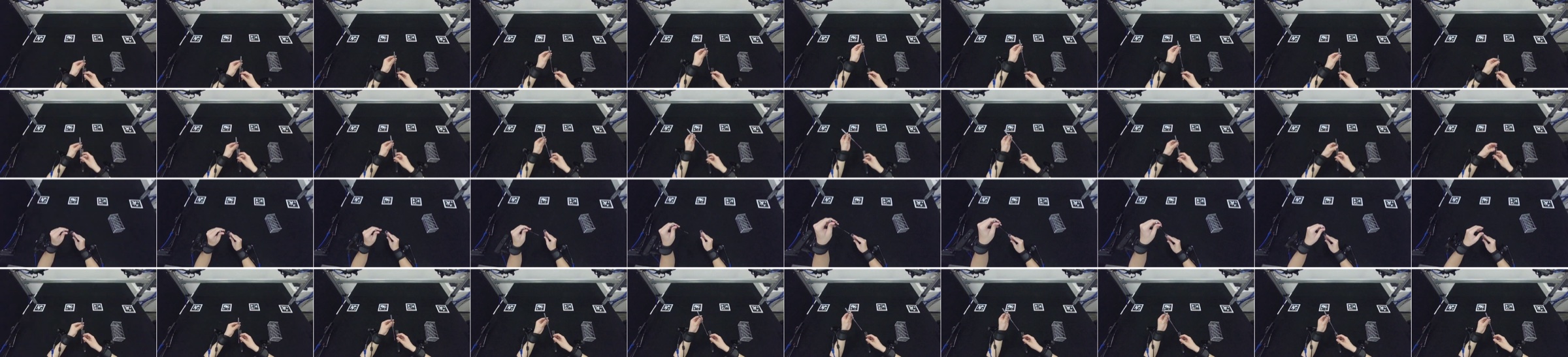}
\caption{Human recovery collection for cup-brush insertion (Task 1).}
\label{fig:collection_t1_hand_recovery}
\end{figure}

\begin{figure}[H]
\centering
\includegraphics[width=\textwidth]{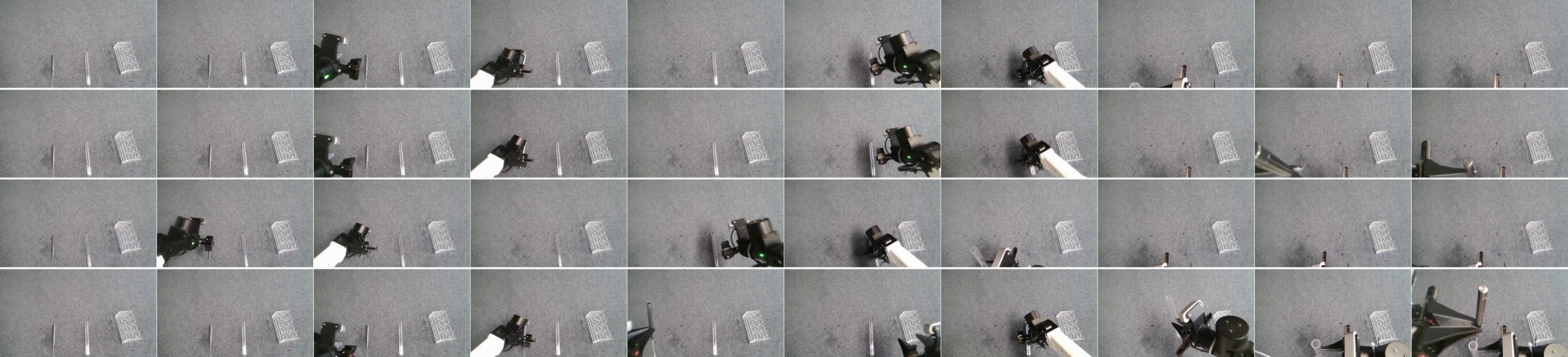}
\caption{Robot success collection for cup-brush insertion (Task 1).}
\label{fig:collection_t1_robot_success}
\end{figure}

\begin{figure}[H]
\centering
\includegraphics[width=\textwidth]{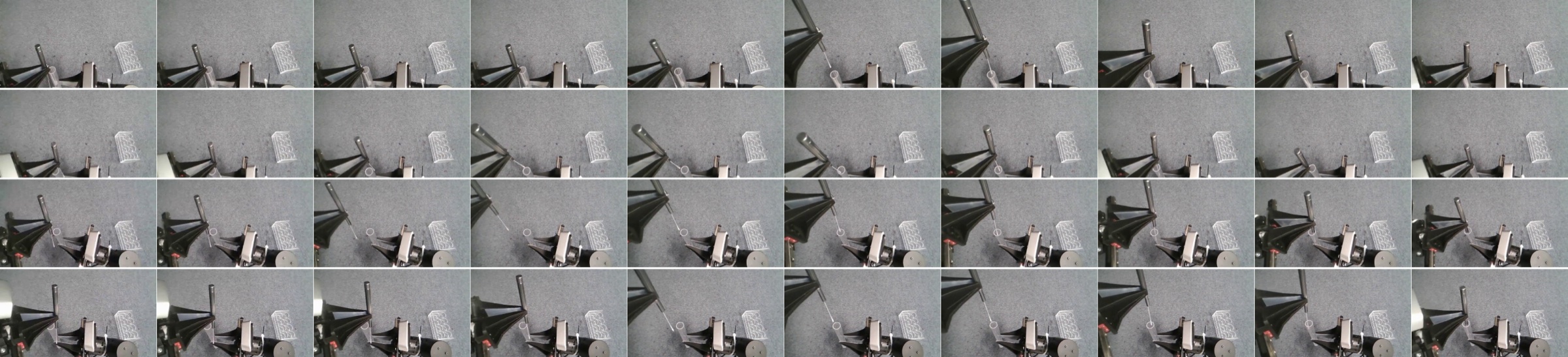}
\caption{Robot recovery collection for cup-brush insertion (Task 1).}
\label{fig:collection_t1_robot_recovery}
\end{figure}

\begin{figure}[H]
\centering
\includegraphics[width=\textwidth]{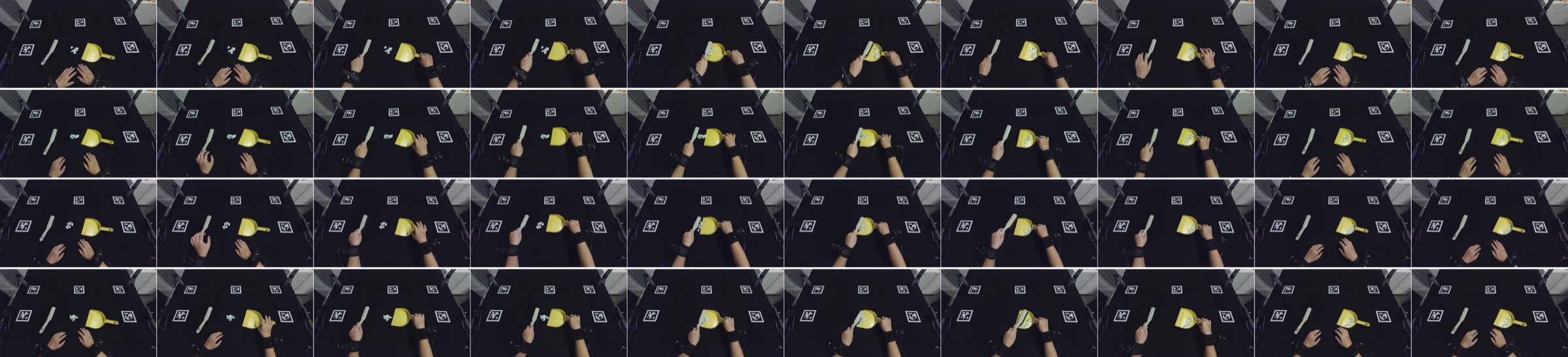}
\caption{Human success collection for table sweep (Task 2).}
\label{fig:collection_t2_hand_success}
\end{figure}

\begin{figure}[H]
\centering
\includegraphics[width=\textwidth]{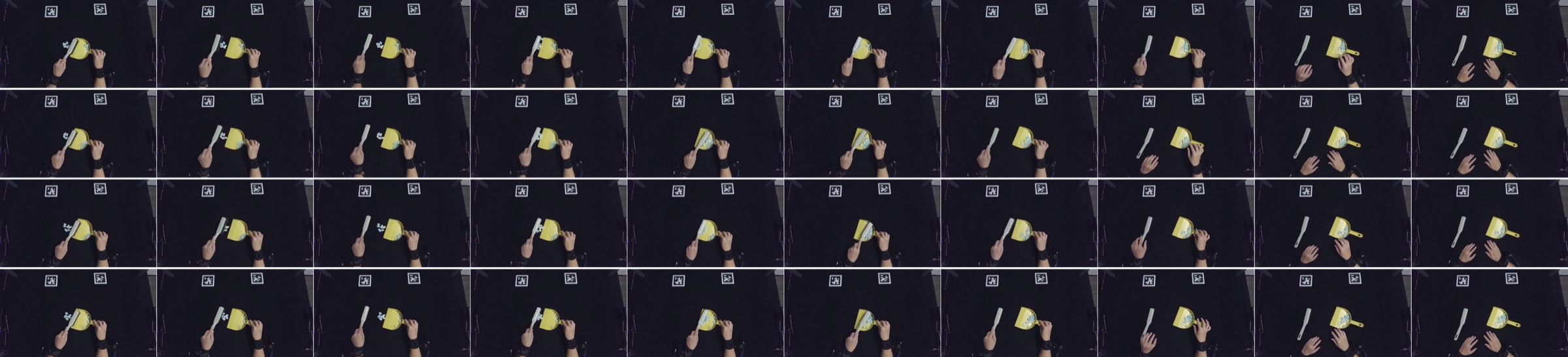}
\caption{Human recovery collection for table sweep (Task 2).}
\label{fig:collection_t2_hand_recovery}
\end{figure}

\begin{figure}[H]
\centering
\includegraphics[width=\textwidth]{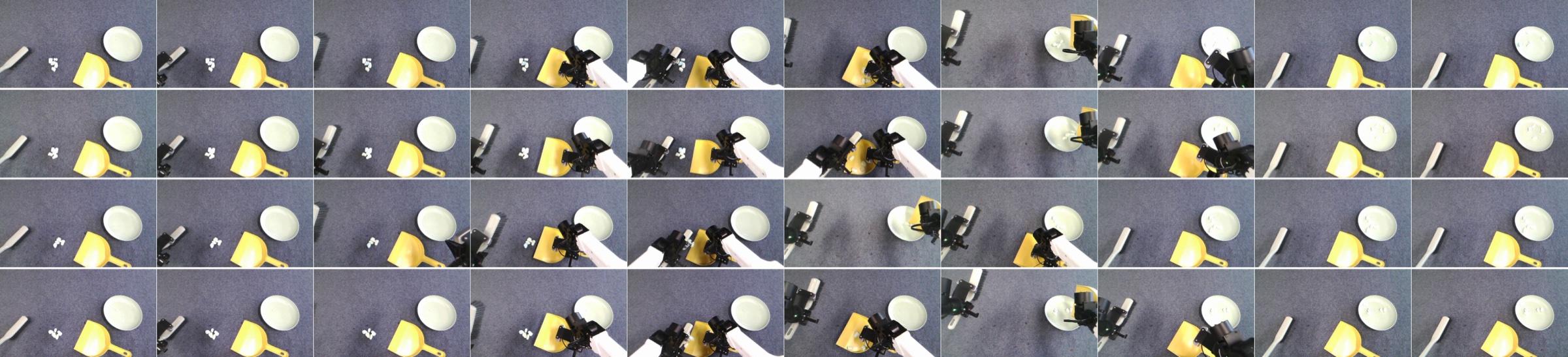}
\caption{Robot success collection for table sweep (Task 2).}
\label{fig:collection_t2_robot_success}
\end{figure}

\begin{figure}[H]
\centering
\includegraphics[width=\textwidth]{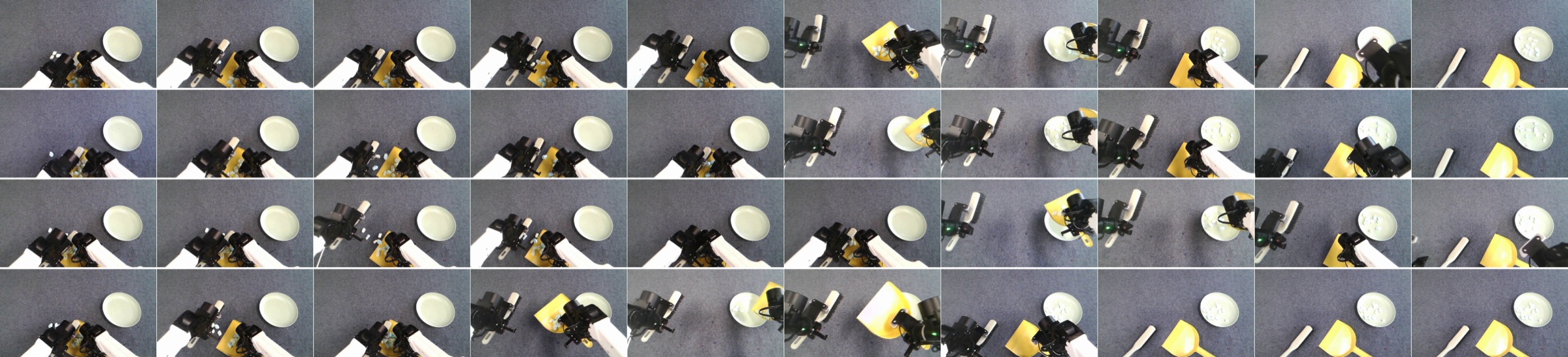}
\caption{Robot recovery collection for table sweep (Task 2).}
\label{fig:collection_t2_robot_recovery}
\end{figure}

\begin{figure}[H]
\centering
\includegraphics[width=\textwidth]{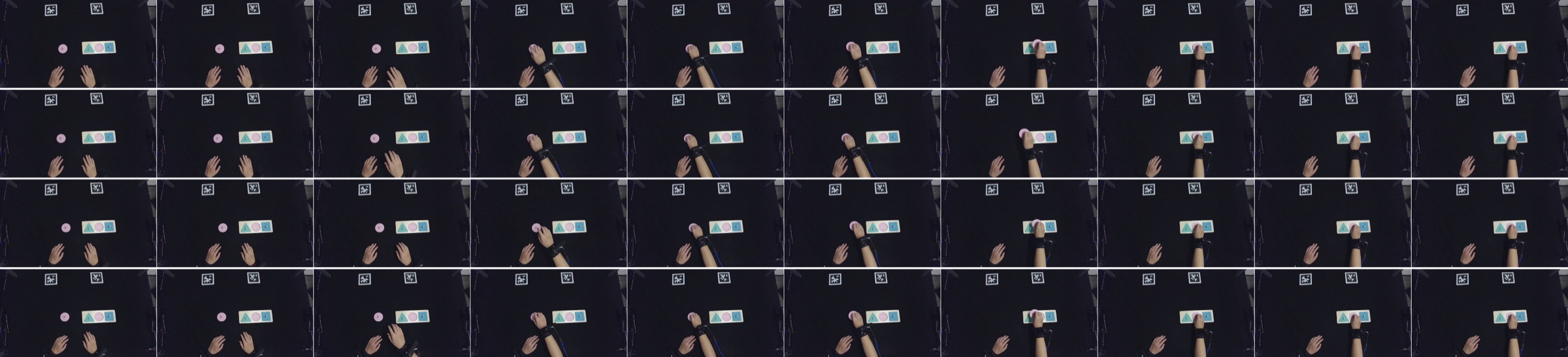}
\caption{Human success collection for round-disk placement (Task 3).}
\label{fig:collection_t3_hand_success}
\end{figure}

\begin{figure}[H]
\centering
\includegraphics[width=\textwidth]{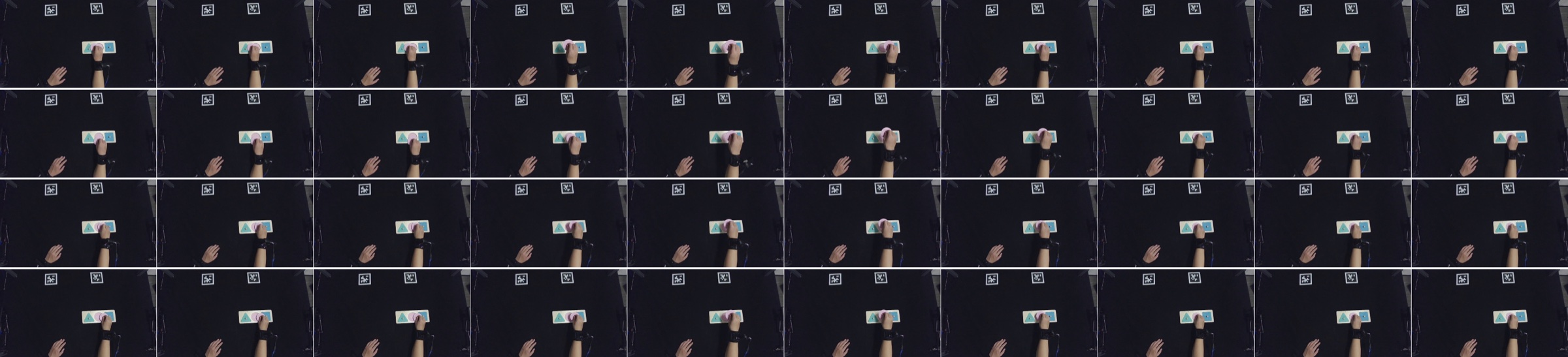}
\caption{Human recovery collection for round-disk placement (Task 3).}
\label{fig:collection_t3_hand_recovery}
\end{figure}

\begin{figure}[H]
\centering
\includegraphics[width=\textwidth]{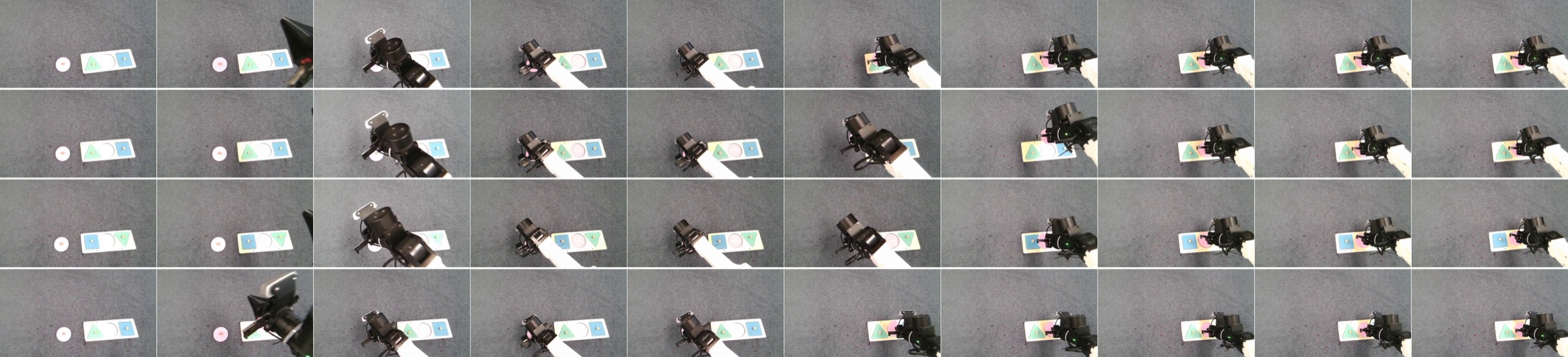}
\caption{Robot success collection for round-disk placement (Task 3).}
\label{fig:collection_t3_robot_success}
\end{figure}

\begin{figure}[H]
\centering
\includegraphics[width=\textwidth]{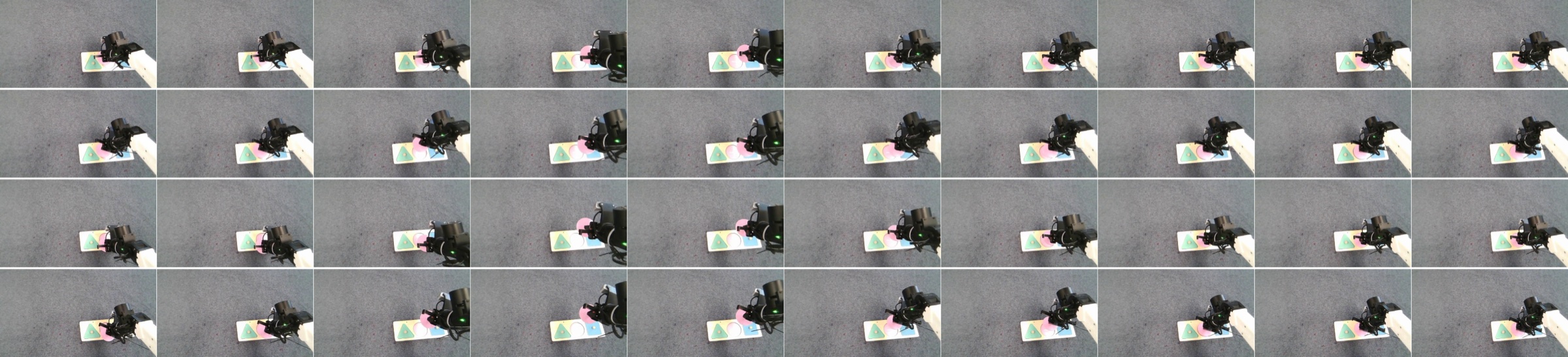}
\caption{Robot recovery collection for round-disk placement (Task 3).}
\label{fig:collection_t3_robot_recovery}
\end{figure}

\begin{figure}[H]
\centering
\includegraphics[width=\textwidth]{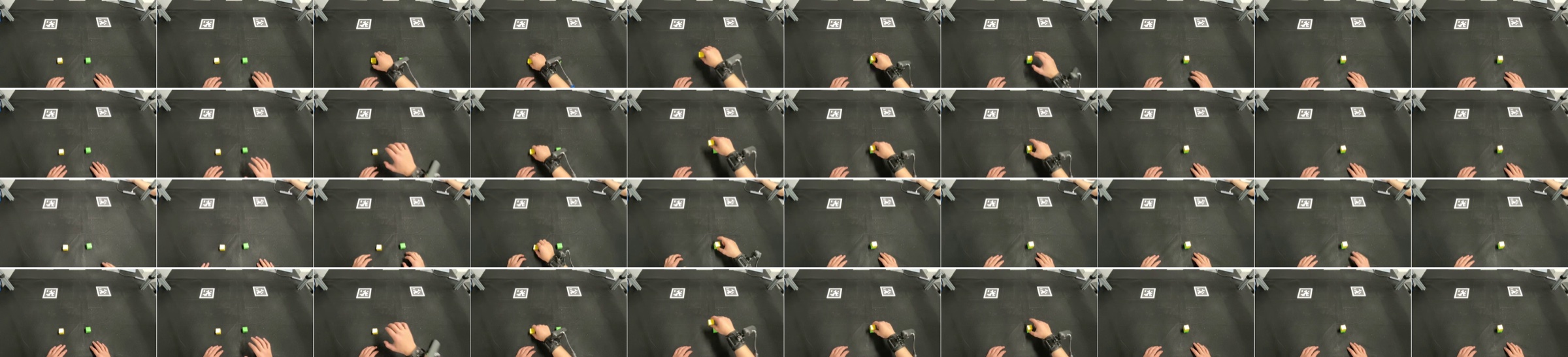}
\caption{Human success collection for cube stacking (Task 4).}
\label{fig:collection_t4_hand_success}
\end{figure}

\begin{figure}[H]
\centering
\includegraphics[width=\textwidth]{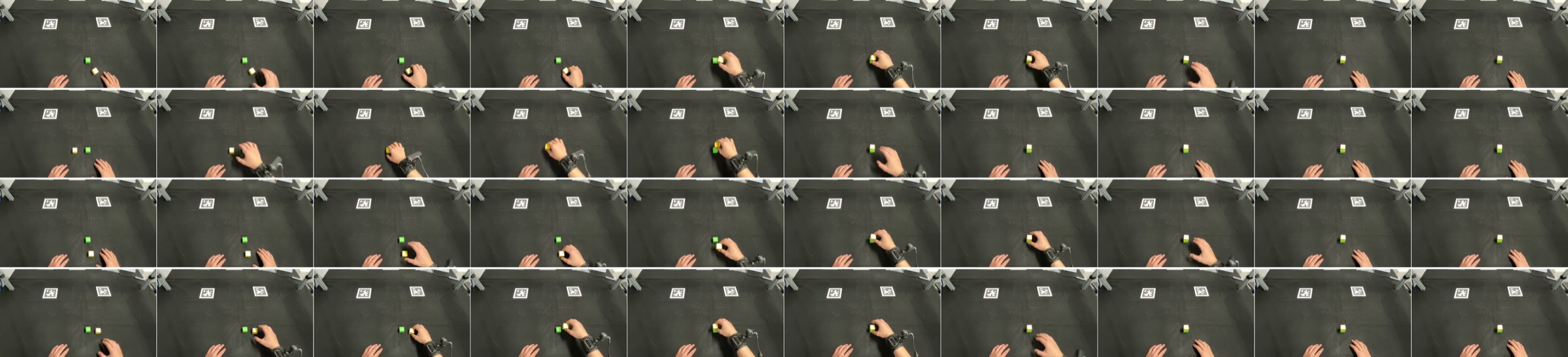}
\caption{Human recovery collection for cube stacking (Task 4).}
\label{fig:collection_t4_hand_recovery}
\end{figure}

\begin{figure}[H]
\centering
\includegraphics[width=\textwidth]{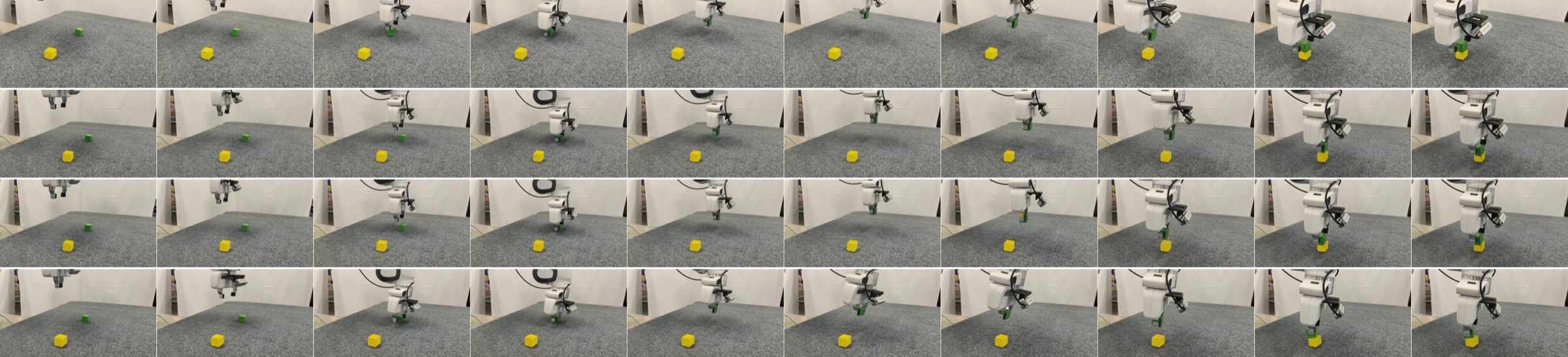}
\caption{Robot success collection for cube stacking (Task 4).}
\label{fig:collection_t4_robot_success}
\end{figure}

\begin{figure}[H]
\centering
\includegraphics[width=\textwidth]{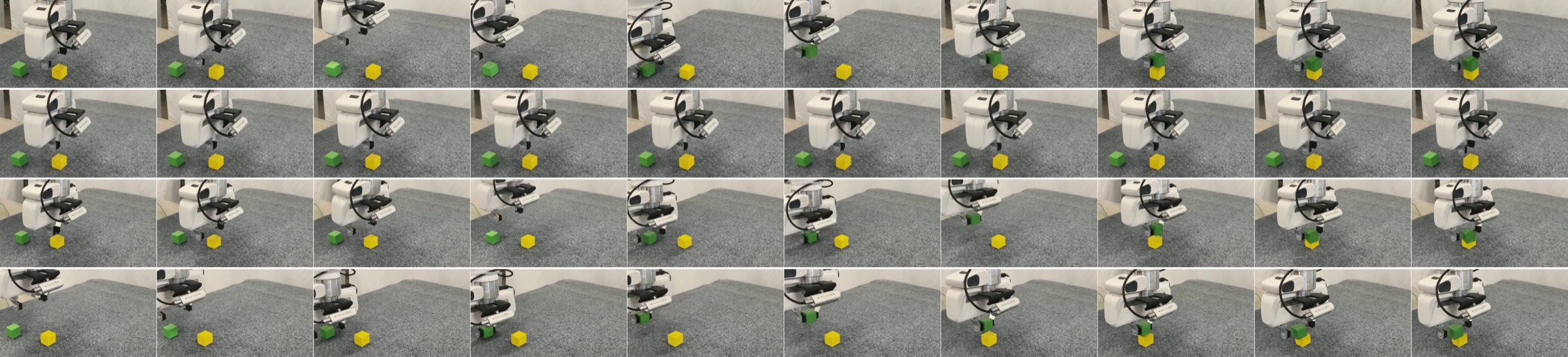}
\caption{Robot recovery collection for cube stacking (Task 4).}
\label{fig:collection_t4_robot_recovery}
\end{figure}

\subsubsection{\textsc{EgoRecovery} Inference}

\begin{figure}[H]
\centering
\includegraphics[width=\textwidth]{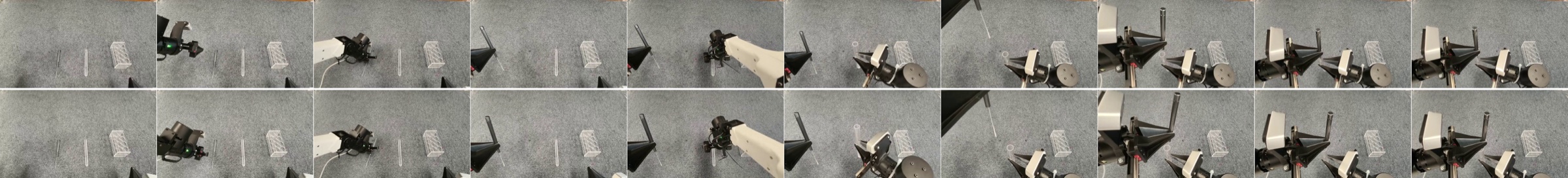}
\caption{\textsc{EgoRecovery} nominal rollout for cup-brush insertion (Task 1).}
\label{fig:inference_t1_robot_success}
\end{figure}

\begin{figure}[H]
\centering
\includegraphics[width=\textwidth]{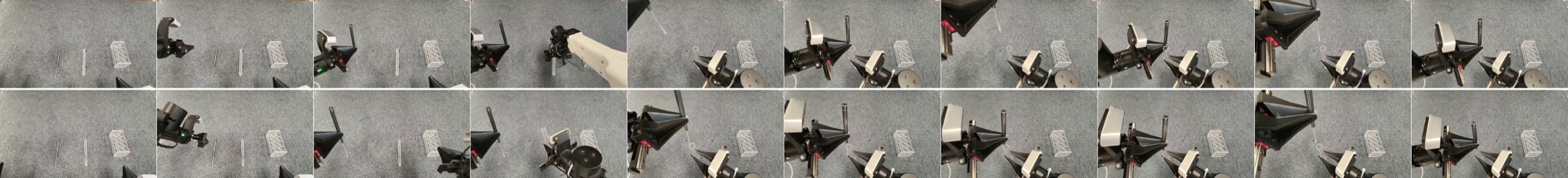}
\caption{\textsc{EgoRecovery} recovery rollout for cup-brush insertion (Task 1).}
\label{fig:inference_t1_robot_recovery}
\end{figure}

\begin{figure}[H]
\centering
\includegraphics[width=\textwidth]{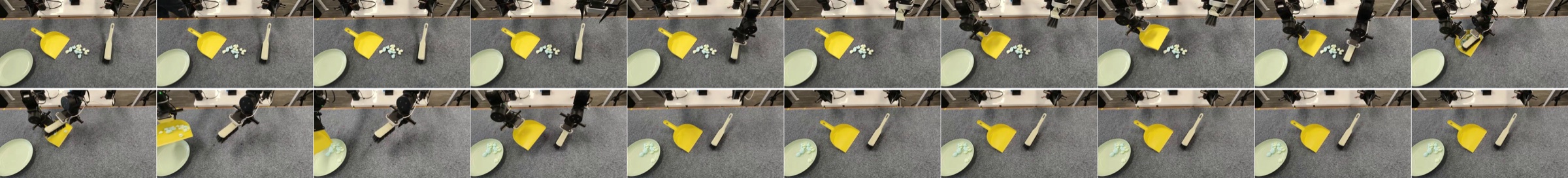}
\caption{\textsc{EgoRecovery} nominal rollout for table sweep (Task 2).}
\label{fig:inference_t2_robot_success}
\end{figure}

\begin{figure}[H]
\centering
\includegraphics[width=\textwidth]{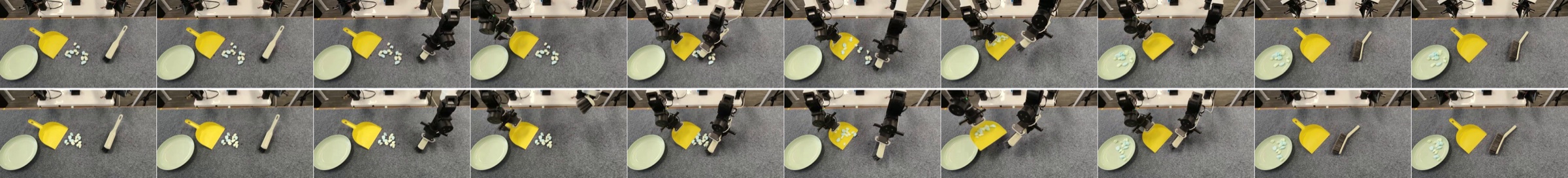}
\caption{\textsc{EgoRecovery} recovery rollout for table sweep (Task 2).}
\label{fig:inference_t2_robot_recovery}
\end{figure}

\begin{figure}[H]
\centering
\includegraphics[width=\textwidth]{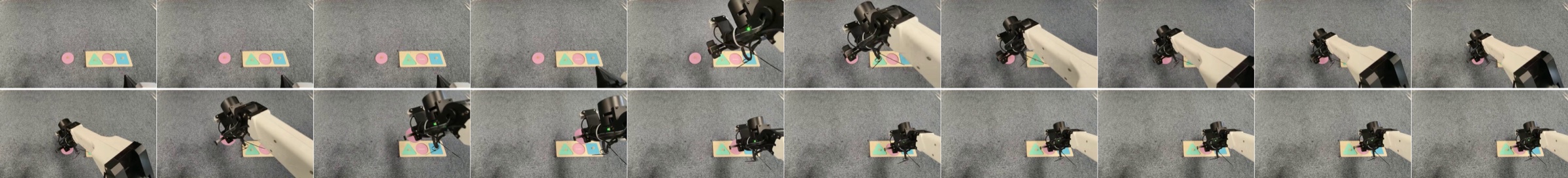}
\caption{\textsc{EgoRecovery} nominal rollout for round-disk placement (Task 3).}
\label{fig:inference_t3_robot_success}
\end{figure}

\begin{figure}[H]
\centering
\includegraphics[width=\textwidth]{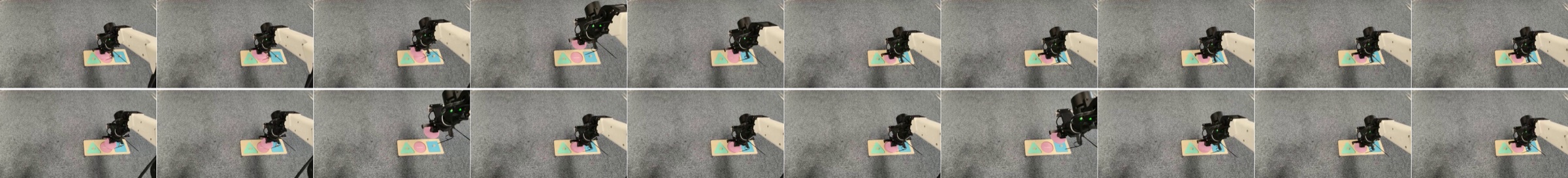}
\caption{\textsc{EgoRecovery} recovery rollout for round-disk placement (Task 3).}
\label{fig:inference_t3_robot_recovery}
\end{figure}

\begin{figure}[H]
\centering
\includegraphics[width=\textwidth]{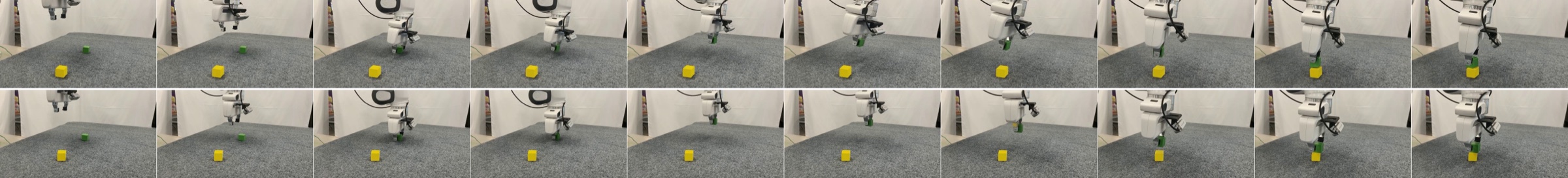}
\caption{\textsc{EgoRecovery} nominal rollout for cube stacking (Task 4).}
\label{fig:inference_t4_robot_success}
\end{figure}

\begin{figure}[H]
\centering
\includegraphics[width=\textwidth]{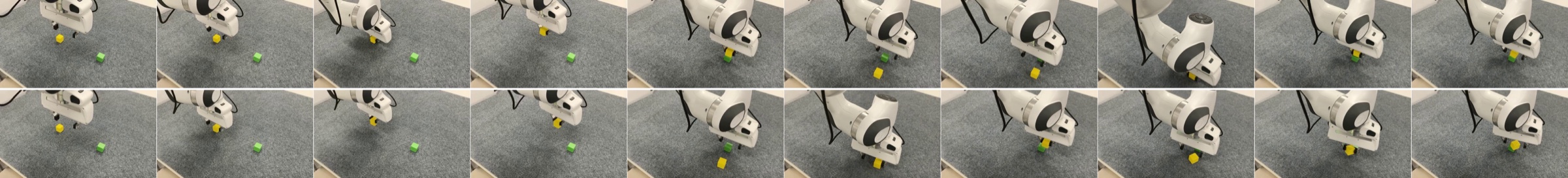}
\caption{\textsc{EgoRecovery} recovery rollout for cube stacking (Task 4).}
\label{fig:inference_t4_robot_recovery}
\end{figure}
 
\end{document}